\def\cvprPaperID{3363}
\def\confName{CVPR}
\def\confYear{2024}

\def\paperTitle{Guess The Unseen: Dynamic 3D Scene Reconstruction from Partial 2D Glimpses}

\def\authorBlock{
    Inhee Lee \qquad
    Byungjun Kim \qquad
    Hanbyul Joo \\
    Seoul National University \\
    {\tt\small \{ininin0516, byungjun.kim, hbjoo\}@snu.ac.kr} \\
    {\tt\small \href{https://snuvclab.github.io/gtu/}{\color{magenta}{https://snuvclab.github.io/gtu/}}}
}

\newif\ifreview 
\newif\ifarxiv \newcommand{\arxiv}{\arxivtrue}
\newif\ifcamera 
\newif\ifrebuttal

\ifcamera \usepackage[accsupp]{axessibility} \fi
\ifcamera  \fi

\arxiv

\pdfoutput=1
\documentclass[10pt,twocolumn,letterpaper]{article}
\ifreview \usepackage[review]{cvpr} \fi
\ifarxiv \usepackage[pagenumbers]{cvpr} \fi
\ifrebuttal \usepackage[rebuttal]{cvpr} \fi
\ifcamera \usepackage{cvpr} \fi

\usepackage{graphicx}
\usepackage{amsmath}
\usepackage{amssymb}
\usepackage{booktabs}

\usepackage{times}
\usepackage{microtype}
\usepackage{epsfig}
\usepackage[table,xcdraw]{xcolor}
\usepackage{caption}
\usepackage{float}
\usepackage{placeins}
\usepackage{color, colortbl}
\usepackage{stfloats}
\usepackage{enumitem}
\usepackage{tabularx}
\usepackage{xstring}
\usepackage{multirow}
\usepackage{xspace}
\usepackage{cuted} %
\usepackage{url}
\usepackage{subcaption}
\usepackage{xcolor}
\usepackage[hang,flushmargin]{footmisc}
\usepackage{bm}
\usepackage{comment}
\usepackage{algorithm} %
\usepackage{algpseudocode} %

\ifcamera \usepackage[accsupp]{axessibility} \fi

\ifarxiv  \fi

\newcommand{\figref}[1]{Fig.~\ref{#1}}
\newcommand{\tabref}[1]{Tab.~\ref{#1}}

\newcommand{\secref}[1]{Sec.~\ref{#1}}

\renewcommand{\vec}[1]{\bm{#1}}
\newcommand{\mat}[1]{\bm{#1}}

\newcommand{\R}[1]{{%
    \textbf{%
        \ifstrequal{#1}{1}{\textcolor{red}{R#1}}{%
        \ifstrequal{#1}{2}{\textcolor{blue}{R#1}}{%
        \ifstrequal{#1}{3}{\textcolor{magenta}{R#1}}{%
        \ifstrequal{#1}{4}{\textcolor{teal}{R#1}}{%
                           \textcolor{cyan}{R#1}%
        }}}}%
    }%
}}

\definecolor{tabfirst}{rgb}{1, 0.7, 0.7} %
\definecolor{tabsecond}{rgb}{1, 0.85, 0.7} %
\definecolor{tabthird}{rgb}{1, 1, 0.7} %

\algnewcommand{\Inputs}[1]{%
  \State \textbf{Inputs:}
  \Statex \hspace*{\algorithmicindent}\parbox[t]{.8\linewidth}{\raggedright #1}
}
\algnewcommand{\Initialize}[1]{%
  \State \textbf{Initialize:}
  \Statex \hspace*{\algorithmicindent}\parbox[t]{.8\linewidth}{\raggedright #1}
}

\usepackage{xr-hyper}

\makeatletter
\newcommand*{\addFileDependency}[1]{
  \typeout{(#1)}
  \@addtofilelist{#1}
  \IfFileExists{#1}{}{\typeout{No file #1.}}
}

\makeatother

\usepackage[pagebackref,breaklinks,colorlinks]{hyperref}
\usepackage[capitalize]{cleveref}
\crefname{section}{Sec.}{Secs.}
\crefname{table}{Table}{Tables}
\crefname{figure}{Fig.}{Figs.}

\begin{document}
\title{\paperTitle}
\author{\authorBlock}
\maketitle

\begin{strip}\centering
\includegraphics[width=\linewidth, trim={0.5cm 7.5cm 0cm 8cm},clip]{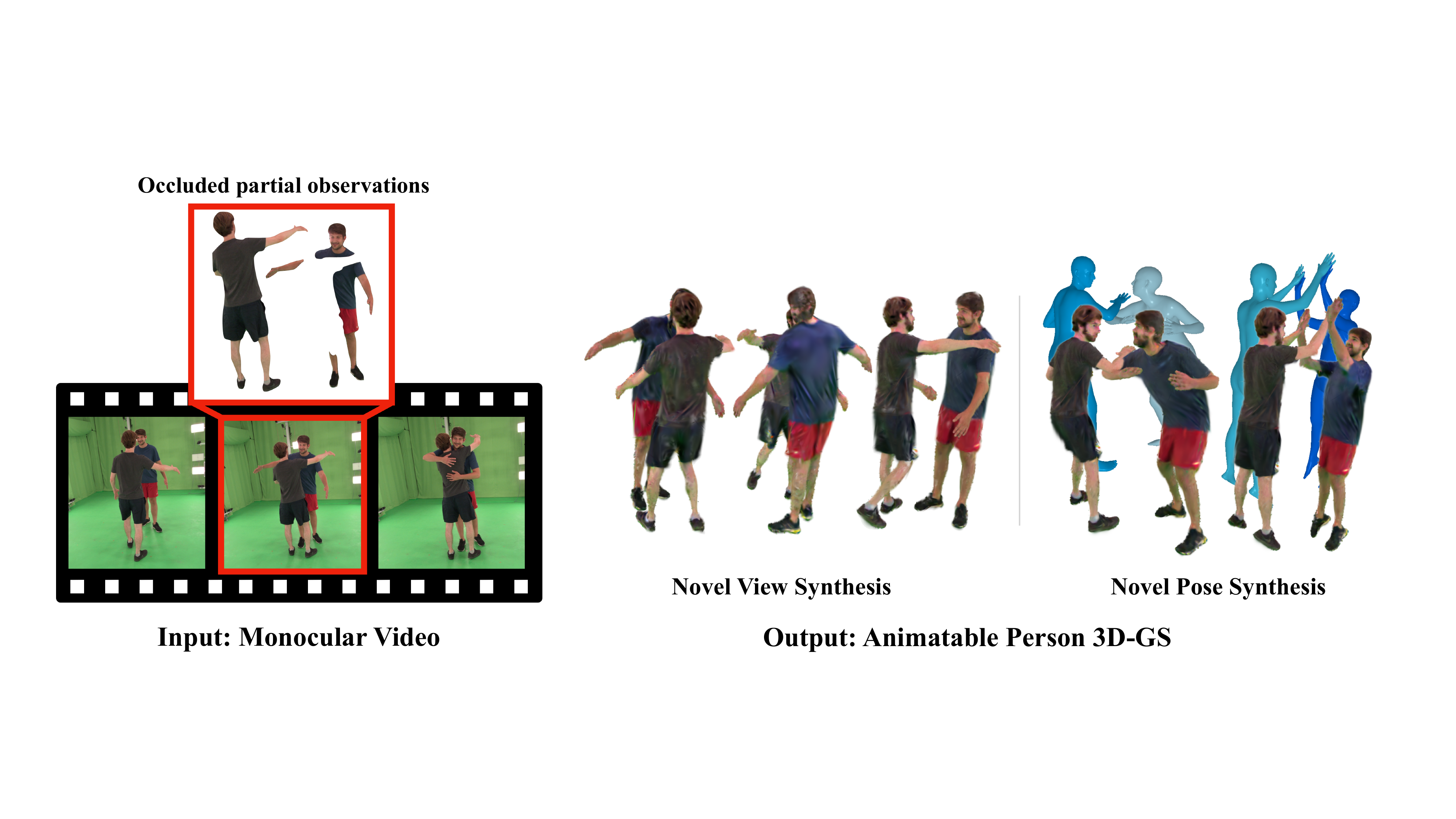}
\vspace{-10pt}
\captionof{figure}{We present a method to reconstruct dynamic scenes from a monocular video capturing partial 2D observations. As a key advantage, our method can estimate the unseen body parts by leveraging a pre-trained diffusion model~\cite{rombach2022stablediffusion_cvpr} via SDS method~\cite{poole2022dreamfusion}. The reconstructed scenes can be rendered to any viewpoint and each human body can be transformed into any body posture controlled by SMPL~\cite{smpl2015} parameters.}
\label{fig:teaser}
\end{strip}

\begin{abstract}
In this paper, we present a method to reconstruct the world and multiple dynamic humans in 3D from a monocular video input. As a key idea, we represent both the world and multiple humans via the recently emerging 3D Gaussian Splatting (3D-GS) representation, enabling to conveniently and efficiently compose and render them together. 
In particular, we address the scenarios with severely limited and sparse observations in 3D human reconstruction, a common challenge encountered in the real world. 
To tackle this challenge, we introduce a novel approach to optimize the 3D-GS representation in a canonical space by fusing the sparse cues in the common space, where we leverage a pre-trained 2D diffusion model to \textbf{synthesize} unseen views while keeping the consistency with the observed 2D appearances. We demonstrate our method can reconstruct high-quality animatable 3D humans in various challenging examples, in the presence of occlusion, image crops, few-shot, and extremely sparse observations.
After reconstruction, our method is capable of not only rendering the scene in any novel views at arbitrary time instances, but also editing the 3D scene by removing individual humans or applying different motions for each human.
Through various experiments, we demonstrate the quality and efficiency of our methods over alternative existing approaches. 
\end{abstract}

\section{Introduction}
\label{sec:intro}

The process of digitizing our world in 3D necessitates the reconstruction of not only static environmental elements but also dynamic objects, particularly humans. Given the limited availability of multi-view camera setups, a groundbreaking leap can be achieved by developing a 4D reconstruction method capable of rendering the scenes at novel views and 
arbitrary times just using a monocular video input. 
Reconstructing static components (e.g., buildings) from monocular video benefits from the well-established multi-view geometry principles~\cite{hz_mvgeometry}, where epipolar geometry constraints across different views still hold at different times. Recent advances have further enhanced the quality of these reconstructions by leveraging implicit 3D representations, as demonstrated by NeRF~\cite{mildenhall2020nerf}, NeuS~\cite{wang2021neus}, and Gaussian Splatting (3D-GS)~\cite{kerbl20233Dgaussians}, resulting in more realistic renderings.

However, the same approach is not directly applicable to dynamically moving components, specifically humans. Early work addresses this problem within the context of general non-rigid structure-from-motion approaches~\cite{Akhter2012BilinearBasis, hspark2010nrsfm}. More recent breakthroughs leverage human prior models, such as SMPL~\cite{smpl2015, Hongyi2020GHUM}, as a way to canonicalize the non-rigid observations from multiple views of the monocular video and transform back into the posed spaces~\cite{chen2023fastsnarf, chen2021snarf, guo2023vid2avatar, jiang2023instantavatar, Weng2022humannerf}. 

Yet, these approaches often assume the scenarios where the camera focuses on the human subject, capturing their entire body while the target person revolves around the camera's field of view. While this approach is suitable for intentionally digitizing a specific individual, they encounter substantial challenges in in-the-wild video scenarios, where humans are captured in partial, occluded, cropped, and sparse observed conditions. See the examples shown in \figref{fig:teaser} and \figref{fig:panoptic_result}. Moreover, reconstructing and rendering multiple individuals along with 3D backgrounds within the existing approaches is non-trivial, mainly due to the complexities of integrating multiple neural radiance field models~\cite{zhang2021editable_stnerf, Julian2021NSG}.

In this paper, we present a method to reconstruct both the static world and multiple dynamically moving humans in 3D from a monocular video input, specifically focusing on scenarios with extremely limited and sparse observations. 
To address this challenge, we represent both the world and multiple humans in a common Gaussian splatting 3D representation~\cite{kerbl20233Dgaussians}. Our approach allows to efficiently compose them for novel view rendering or scene editing. 
Notably, to tackle the scenarios with extremely limited and sparse observations in 3D human reconstruction, we introduce a novel approach to optimize the 3D GS representation in a canonical space by fusing the sparse cues in the common space. 
As a core idea, we leverage a pre-trained 2D diffusion model, equipped with Texture Inversion~\cite{gal2022textual_inversion}, to synthesize unseen views without losing the consistency with the observed 2D appearances~\cite{rombach2022stablediffusion_cvpr, poole2022dreamfusion}. 
Via thorough evaluations, we demonstrate that our approach successfully reconstructs high-quality animatable 3D human avatars of dynamically moving individuals from sparse and partial observations. The animatable nature of our 3D human reconstruction outputs enables us to replay the observed motions of the humans in novel views and edit the postures of the humans with arbitrary motions as well. 
Our contribution is summarized as:
(1) representing both a 3D world and multiple humans in a common 3D GS representation for efficient composing and rendering; (2) reconstructing and canonicalizing animatable 3D humans from sparse and partial 2D observations by incorporating SDS loss with a diffusion model and textual inversion; and (3) introducing efficient 4D scene reconstruction and editing pipeline.

\section{Related Work}
\label{sec:related}
\noindent \textbf{Human Reconstruction from Monocular Video.}
 There has been a series of approaches~\cite{xu2018MonoPerfCap,jiang2022selfrecon, peng2021neuralbody_zjumocap, Weng2022humannerf, yu2023monohuman, jiang2022neuman, yu2023monohuman} to reconstruct 3D human avatars from a monocular video capturing moving humans. 
 Mostly, they tackle the problem by finding the correspondences between each frame and optimizing them in a common canonical space.
 To find the correspondences across the frames, diverse prior knowledge is leveraged such as parametric model~\cite{smpl2015, Alldieck2018people_snaphost}, pixel-aligned features~\cite{Neverova2020cse}, or optical flow~\cite{yang2022banmo}. After the success of NeRF~\cite{mildenhall2020nerf}, recent methods~\cite{jiang2022selfrecon, peng2021neuralbody_zjumocap, Weng2022humannerf, yu2023monohuman} use NeRF and its variants to reconstruct a human by leveraging a parametric model SMPL~\cite{smpl2015}. 
  HumanNeRF~\cite{Weng2022humannerf} and SelfRecon~\cite{jiang2022selfrecon} improve the reconstruction quality by correcting the inaccurate canonicalization originated by non-rigid deformation.
Vid2Avatar~\cite{guo2023vid2avatar} and Neuman~\cite{jiang2022neuman} reconstruct a human without a mask by learning a background jointly. InstantAvatar~\cite{jiang2023instantavatar} reduces the required time of optimization from a few hours into a minute leveraging iNGP~\cite{mueller2022instantngp}. OccNeRF~\cite{xiang2023occnerf} proposes a method that can reconstruct people even with occlusion, using surface-based rendering and visibility attention.
However, unlike our method, all of the above except OccNeRF assume the person is not occluded and most of the body is shown in the video, which is rare in in-the-wild videos. 
 
\noindent \textbf{Diffusion on 3D Tasks.}
After the recent breakthroughs of diffusion models on image generation task~\cite{ho2020ddpm}, several methods suggest a way to use diffusion model on 3D tasks~\cite{poole2022dreamfusion, wang2023sjc, liu2023zero123, zhou2023sparsefusion, deng2023nerdi,lei2023rgbd2,shao2022diffustereo}. For example, RGBD2~\cite{lei2023rgbd2} trains an RGB-D diffusion model to complete the unobserved area of a room using diffusion inpainting approach~\cite{lugmayr2022repaint} and DiffuStereo~\cite{shao2022diffustereo} trains diffusion-stereo network for higher reconstruction quality in sparse multi-view settings. 
In particular, the SDS~\cite{poole2022dreamfusion} method which leverages a pretrained text-to-image diffusion model~\cite{saharia2022photorealistic} has been applied widely, such as text-to-3D~\cite{poole2022dreamfusion, wang2023sjc, wang2023prolificdreamer} or single image-to-3D generation tasks~\cite{liu2023zero123, lin2023magic3d, melas2023realfusion}. However, the vanilla SDS loss is not 3D consistent itself and prone to artifacts like the Janus effect. To improve the SDS loss, many techniques are proposed such as leveraging 3D prior~\cite{liu2023zero123}, fine-tuning diffusion~\cite{wang2023prolificdreamer}, giving better conditions~\cite{melas2023realfusion} and using advanced optimization schemes~\cite{lin2023magic3d, Chan2023iccv_GeNVS, Chen_2023iccv_fantasia3d, Chen_2023iccv_fantasia3d}. 
SDS loss is also applied to make better 3D avatars recently~\cite{liao2023tada_3dv_accepted, huang2024tech}. However, there has been no trial on completing an imperfect reconstruction of a human with a diffusion. 

\noindent \textbf{Compositional Human-Scene Reconstruction.}
 Separating 4D scenes into static backgrounds and dynamic objects is a common approach in the 4D reconstruction problem. The static-dynamic separation can be done by prior knowledge on the targets~\cite{Julian2021NSG, Abhijit2022PNF}, such as cars and pedestrians are dynamic, or can be performed automatically by minimizing a certain energy term~\cite{wu2022d2nerf}.  
 Recent human reconstruction methods also use compositional approaches to reconstruct a human from videos~\cite{guo2023vid2avatar,jiang2022neuman, Shuai2022soccer_dataset, zhang2021editable_stnerf}. For monocular reconstruction, Neuman~\cite{jiang2022neuman} and Vid2avatar~\cite{guo2023vid2avatar} use two separate implicit functions each of which represents background and person respectively. While they allow the model to reconstruct a person regardless of the person's mask quality, these models cannot handle occlusion and multiple people at once.
 In a multi-view setting, \emph{Shuai et al.}~\cite{Shuai2022soccer_dataset} tackles more practical situations where multiple people interact with objects. It models each person, object, and background with a NeuralBody~\cite{peng2021neuralbody_zjumocap} and NeRF~\cite{mildenhall2020nerf} respectively and renders them compositionally through ST-NeRF~\cite{zhang2021editable_stnerf} pipeline. However, it requires 8 multi-view videos as input, which is unavailable in in-the-wild situations.

\section{Preliminaries}
\label{sec:preliminaries}
In our method, we use 3D Gaussian Splatting~\cite{kerbl20233Dgaussians} to represent 4D scenes, and use Score Distillation Sampling (SDS) as a tool to estimate unseen human body parts. Here, we provide an overview of these preliminary concepts. 
\subsection{3D Gaussian Splatting}

3D Gaussian Splatting (3D-GS) is an explicit 3D representation to model a radiance field of a static 3D scene with a set of 3D Gaussians and their attributes~\cite{kerbl20233Dgaussians}.
A 3D static scene can be modeled by a set of 3D Gaussians $\{G_i\}_{i=1}^{M}$ where the $i$-th Gaussian is represented by  
$G_i\ = \{\vec{\mu}_i,  \vec{q}_i,  \vec{s}_i, \vec{c}_i, o_i\}$, where $\vec{\mu} \in \mathbb{R}^{3}$ is the Gaussian center, $\vec{s} \in \mathbb{R}^3$ and $\vec{q} \in SO(3)$ are respectively the scaling factor and the rotation represented in quaternion to define the covariance matrix $\mat{\Sigma} \in \mathbb{R}^{3\times3}$, $\vec{c_i}\in \mathbb{R}^3$ is the color, and $o_i \in \mathbb{R}$ is opacity.
For a 3D query location $\vec{x}\in \mathbb{R}^3$, its Gaussian weight $\mathbf{g}(\vec{x})$ is represented as:
\begin{equation}
     \mathbf{g}(\vec{x}) = e^{-\frac{1}{2}(\vec{x}-\vec{\mu})^T\mat{\Sigma}^{-1}(\vec{x}-\vec{\mu})},
\end{equation}
where the symmetric 3D covariance matrix $\mat{\Sigma} \in \mathbb{R}^{3\times3}$  is represented by
\begin{align}
    \mat{\Sigma} = \mat{R} \mat{S} \mat{S}^T \mat{R}^T.
 \end{align}
$\mat{R}=\text{quat2rot}(\vec{q})$ is  a rotation matrix converted from $\vec{q}$, and $\mat{S} = \text{diag}(\vec{s})$ is a diagonal matrix from scaling factor $\vec{s}$.

3D-GS rasterizes these 3D Gaussians $\{{G_i}\}_{i=1}^M$ by sorting them in depth order in camera space and projecting them to the image plane. If $N$ number of Gaussians are projected on 2D location $\vec{p} \in \mathbb{R}^2$, the pixel color $C(\vec{p})$ is given by $\alpha$-blended rendering as follows:
\begin{align}
    \label{eq:3dgs_rendering}
    C(\vec{p}) &= \sum_{i\in N} c_i \alpha_i \prod_{j=1}^{i-1} (1-\alpha_j), \\
    \alpha_i &= \mathbf{g}^{2D}_{i}(\vec{p}) \cdot o_i,
\end{align}
where $\mathbf{g}^{2D}_i$ is the weight after the 2D projection of 3D Gaussian $\mathbf{g}_{i}$ to the image plane, and we use the Jacobian of the affine approximation of the projective transformation, following previous approaches~\cite{3DGS_gaussian_projection_theory, kerbl20233Dgaussians}.
As the output of 3D scene reconstruction, we obtain the parameters of 3D Gaussians $\mathcal{G} = \{G_i\}_{i=1}^{M}$ by optimizing them with reconstruction loss which is calculated from the rendering~\cref{eq:3dgs_rendering}.

\subsection{Score Distillation Sampling}
 Score Distillation Sampling (SDS) method~\cite{poole2022dreamfusion} is an approach that leverages the prior knowledge underlying text-to-image (T2I) diffusion models to generate 3D content. SDS optimizes any differentiable 3D representation $\vec{\Theta}$ by aligning rendered output $\{\vec{I}\}$ from arbitrary views to be on the distribution of diffusion model $\phi$.
 This can be achieved by minimizing the residual between noise $\vec{\epsilon}$, which perturbs $\vec{z}$ into $\vec{z}_\tau$, and predicted noise $\vec{\epsilon}(\vec{z}_\tau; y,\tau)$ where $\vec{z}$ is a latent of $\{\vec{I}\}$ encoded by VAE of latent diffusion model~\cite{rombach2022LDM}. By omitting gradient through diffusion model $\phi$, the SDS loss can be written as follows:
\begin{equation}
    \nabla_{\Theta}\mathcal{L}_{SDS}= \mathbb{E}_{t,\vec{\epsilon}}\left[ 
    \omega(\tau)( \vec{\epsilon}_{\phi}(\vec{z}_{\tau} ; y, \tau) - \vec{\epsilon} ) 
     \frac{\partial \vec{z} }{\partial \vec{I}} \frac{\partial \vec{I}}{\partial \Theta }
    \right],
\end{equation}
where $\vec{z}_\tau$ is a noised latent with time step $\tau$ and $y$ is text prompt conditioning diffusion model $\phi$. The $\omega(\tau)$ denotes the weighting function defined by the scheduler of the diffusion model.

\begin{figure}[t]
    \centering
     \includegraphics[width=1.0\columnwidth, trim={0 0 -2 0},clip]{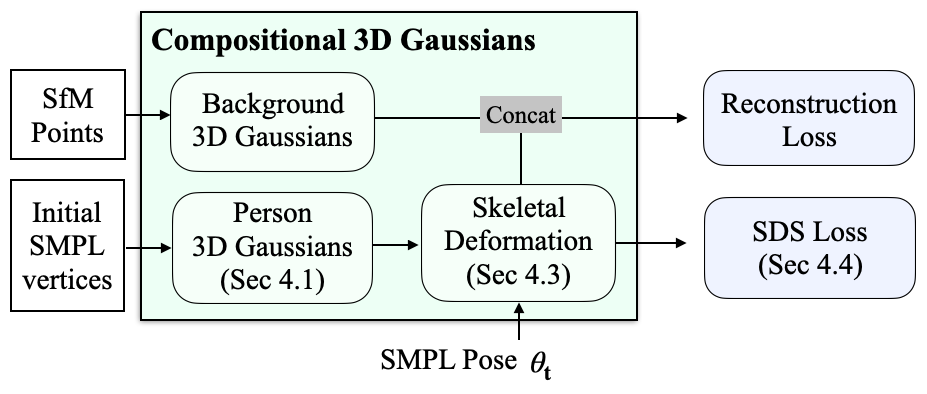}
     \vspace{-20pt}
    \caption{\textbf{Method overview.} Overview of our pipeline. (\secref{sec:method}).}
    \label{fig:overview}
\vspace{-5pt}
\end{figure}

\section{Method}
\label{sec:method}
\subsection{Overview}

Our model reconstructs dynamically moving multi-humans and the static background jointly from a casually captured monocular video. Our system takes, as input, $T$ frames of images $\{I_t\}_{t=1\dots T}$ with corresponding camera parameters $\{\mathbf{P}_t\}_{t=1\dots T}$, and outputs the 3D scenes in the representation of 3D Gaussian Splatting $\mathcal{G}^{BG}$ and $\{ \mathcal{G}^{h}_{j} \}_{j=1\dots N}$, where $\mathcal{G}^{BG}$ is to represent the 3D background and $\mathcal{G}^{h}_{j}$ is for the $j$-th human. 

Importantly, we represent the individual human in a canonical space mapped to the rest pose (or A-pose) of SMPL model~\cite{smpl2015}, which can be transformed to any ``posed'' space parameterized by SMPL pose parameter $\vec{\theta} \in \mathbb{R}^{72}$. 
Thus, the appearance of the $j$-th human at time $t$ can be represented as $\mathcal{G}^{h}_{j}(\vec{\theta}_{j,t})$ by inputting the corresponding SMPL parameters $\vec{\theta}_{j,t}$ for $j$-th human at time $t$.
Note that within our representation, the posture of each individual can be controlled independently from other scene parts. This provides us the flexibility to edit people's body motions using various available motion capture data~\cite{mahmood2019amass}.

Since we build both dynamic humans and backgrounds in the same 3D-GS representation, we can effectively render the whole scene by compositing 3D-GS representations, $\mathcal{G}^{\text{All}} = \{ \mathcal{G}^{BG} \} \cup \{ \mathcal{G}^{h}_{j} \}_{j=1\dots N} $, where we can use the same rendering function \cref{eq:3dgs_rendering} without any modification. This shows the major advantage over the alternative approaches such as NeRF-based representation~\cite{guo2023vid2avatar, Weng2022humannerf}, where compositing multiple humans is not trivial. 
Our method is much more convenient and efficient, showing much faster rendering speed (e.g., 40 times) than the competing approaches as demonstrated in our experiments. 

Notably, we mainly focus on reconstructing the 3D humans from sparse monocular observations, in the presence of severe occlusions, cropped views, and few shots, which are commonly observable in the wild.
We address this challenge by fusing the observed cues into the canonical spaces, for which we introduce the transformation between the posed space to the canonical space, while we leverage 3D-GS representation (\cref{sec:person_gaussian}). As a core contribution, we also present a solution to include 2D diffusion prior as a way to synthesize the missing and unobserved part of target human while keeping the consistency to the observed parts (\cref{sec:diffusion_guide}), where we further enhance the quality by incorporating Texture Inversion technique~\cite{kumari2022customdiffusion} to better preserve the target identity. 
The \figref{fig:overview} shows the overall pipeline of our optimization.

\begin{figure}[t]
    \centering
     \includegraphics[width=1.0\columnwidth, trim={0 0 0 0},clip]{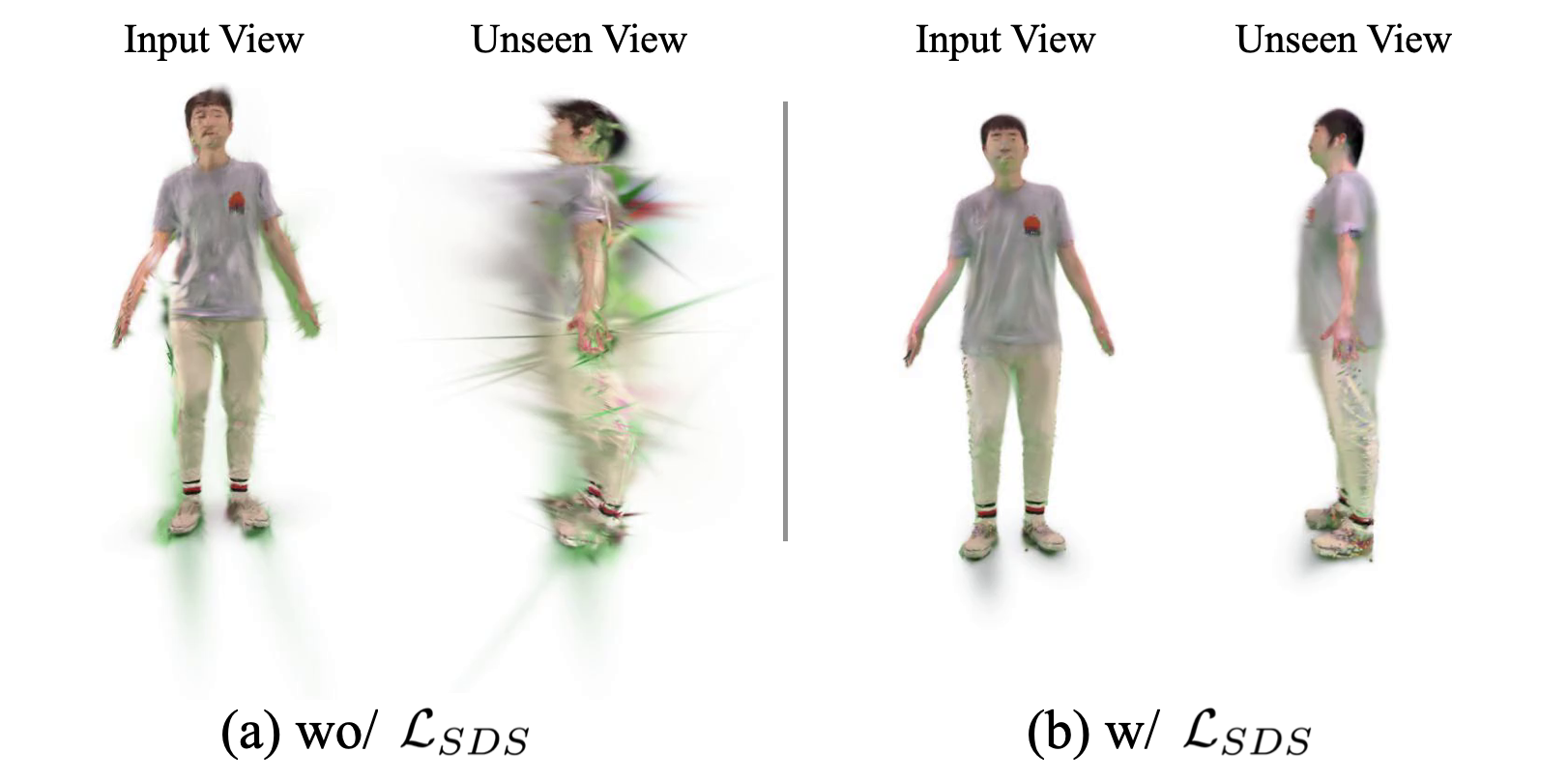}
     \vspace{-15pt}
    \caption{\textbf{Failure examples of optimizing 3D-GS naively.} (a) shows that naively optimizing 3D-GS suffers from artifacts shaped like a hedgehog in unseen view and input view. (b) shows that our SDS loss effectively removes the artifacts observed in both input and unseen views. }
    \label{fig:failure_case}
\vspace{-15pt}
\end{figure}

\subsection{Initializing and Densifying Gaussians}
To represent the background and humans via 3D-GSs, we initialize each representation via the available cues. 
The background Gaussians $\mathcal{G}^{BG}$ are initialized with point cloud obtained by Structure-from-Motion (SfM)~\cite{schoenberger2016sfm_colmap}. In the cases of a fixed camera input where SfM cannot be applicable, we assume that the background is a large 3D sphere with background texture, centered on the mean position of humans.
We represent a $j$-th human via 3D-GS in a canonical space (denoted as subscript $c$), $\mathcal{G}^h_{j,c}$, which is initialized by the vertices of A-posed SMPL mesh $\mathcal{V}(\vec{\beta}_j, \vec{\theta_c})$, where $\vec{\beta}_j$ and $\vec{\theta_c}$ are the SMPL shape and canonical pose parameters respectively, regressed using a monocular 3D pose regressor~\cite{sun2021romp,rajasegaran2022tracking_phalp}. The color and opacity are set in grey and $0.9$ respectively. We assume the SMPL shape parameter $\beta_j$ from the pose regressor is fixed for each human while training.%
 
 To capture the fine details of the background and human, we densify the initial Gaussians adaptively~\cite{kerbl20233Dgaussians} every $N_{den}$ iteration. The Gaussians to be densified are chosen based on the accumulated gradients $\nabla\vec{\mu}_i$, by summing the gradient on the center of Gaussians $\vec{\mu}_i$ computed in each iteration. If the accumulated gradients $\sum_{N_{den}} \nabla \vec{\mu}_i$ is bigger than a predefined threshold, we densify the Gaussian $G_i$ by cloning or splitting it. 
\subsection{Canonicalizing Dynamic Humans}\label{sec:person_gaussian}

To fuse the cues of an individual with different poses across different frames, we model each person with a single canonical model $\mathcal{G}^h_{j, c}$ (for brevity, we drop the person index $j$ in the subscript). 
We also model the deformation function $\mathcal{G}^h_{d}(t) = F_d^{h}(\mathcal{G}^h_{c}, \vec{\theta}_{t})$ which transforms Gaussians in canonical into posed $\mathcal{G}^h_{d}(t)$ at time $t$, following the SMPL pose parameter $\vec{\theta}_{t}$.
Our deformation function $F_d^{h}: \mathbb{R}^3\rightarrow SE(3)$, which maps canonical space into posed space, is defined via the Linear Blend Skinning (LBS) based on the SMPL~\cite{smpl2015}. It translates the center of Gaussian $\vec{x}_i$ and rotates the covariance matrix $\mat{\Sigma}_i$ as follows:
\begin{align}
    \label{eq:skeletal_deformation}
    \vec{x}_{i,p} & = \sum_{k=1}^{N_{joint}} w_k(\vec{x}_{i,c}) (\mat{R}_k\vec{x}_{i,c} + \vec{T}_k), \\
    \mat{\Sigma}_{i,p} & =  \mat{R}_{wei} \mat{\Sigma}_{i,c}\mat{R}_{wei}^T,
\end{align}
where $\mat{R}_k$ and $\vec{T}_k$ are rotation and translation of $k^{th}$ joint of SMPL skeleton which is computed from $\vec{\theta}$ and  $\vec{\beta}_j$. The $\mat{R}_{wei}$ is a derivation of LBS equal to the weighted sum of rotations $\{\mat{R}_k\}_{k=1}^{N_{joint}}$ as follows:
 \begin{equation}
    \label{eq:skeletal_deformation_derivative}
    \mat{R}_{wei} = \sum_{k=1}^{N_{joint}} w_k(\vec{x}_{i,c})\mat{R}_k.
 \end{equation}
 The skinning weight $w_k(\vec{x})$ is calculated from the aligned SMPL vertices defined in canonical space. Here we get $w_k(\vec{x})$ by summing the skinning weight of the $30$ nearest vertices with Inverse Distance Weight (IDW). To accelerate rendering speed, we pre-calculate and store the skinning weight in a voxel grid, similar to SelfRecon~\cite{jiang2022selfrecon}. In each rendering time, we obtain the skinning weight by trilinear interpolation on the weight grid instead of searching the nearest SMPL vertices.

\subsection{Diffusion-Guided Reconstruction} \label{sec:diffusion_guide}

We employ the 2D diffusion prior~\cite{rombach2022stablediffusion_cvpr} in optimizing 3D-GS to represent a human model, as a key idea to overcome sparse observations for the target human in input videos. The intuition behind our approach is that the quality of a 3D human model can be measured by assessing the realism of the rendered images in novel views.  For example, naively optimizing 3D Gaussians $\mathcal{G}$ from sparse and occluded views results in artifacts or missing parts, as shown in \figref{fig:failure_case} (a). The use of the diffusion model, particularly the SDS loss~\cite{poole2022dreamfusion}, can be beneficial to improving the quality of the desired 3D model by enforcing realism in the rendered images at novel views. The SDS loss can be considered as additional virtual cameras guided by the pre-trained 2D diffusion model~\cite{rombach2022stablediffusion_cvpr}.

For each iteration, we make rendering $R^h_v$ of the target human's 3D-GS $\mathcal{G}^{h}$ with a virtual camera $v$ which is randomly sampled from a sphere that is centered on each human and viewing its body.
To give more diversity to the rendering, we also randomly sample the body pose $\theta$ of the person and transform the 3D-GS $\mathcal{G}^{h}$ into the posed space. 
We randomly sample $\vec{\theta}$ either among observed poses or the canonical A-pose($\vec{\theta}_c$), which can be written as $\{\vec{\theta}_t\}_{t\in[1\dots T]} \cup \{\vec{\theta}_c\}$.
With the rendered image $R^{h}_v$ at viewpoint $v$, we compute the SDS loss~\cite{poole2022dreamfusion}, which is proportional to the difference between added noise $\epsilon$ and estimated noise $\epsilon_\phi$ by diffusion model $\phi$: 
 \begin{equation}
     \nabla \mathcal{L}_{SDS} = \mathbb{E}_{t,\epsilon} \Big[ \omega(\tau) (\epsilon_{\phi}(\vec{z}_\tau; \vec{y}, \tau) - \epsilon) \frac{\partial\vec{z}}{\partial{R^{h}_v}}\frac{\partial R^{h}_v}{\partial{\mathcal{G}^{h}}}\Big]
 \end{equation},
where $\vec{z}_\tau$ is a noised latent of rendering $R^{h}_v$, $\tau \in [0,1]$ is a time-step of noise and $\vec{y}$ is conditions applied on the diffusion model. 
The SDS loss mitigates the artifacts in our 3D-GS human model by enforcing the rendered output $R^{h}_v$ in a novel pose to be plausible.

\paragraph{Textual Inversion on SDS.} 
We further improve the efficacy of our SDS method by leveraging the concept of Texture Inversion (TI)~\cite{gal2022textual_inversion}, as a way to make the SDS loss to synthesize the human appearance similar to our target identity, rather than generating arbitrary appearance. 
Applying the SDS loss only with text prompts, such as ``A photo of a person'', may easily converge to a random human appearance due to the diversity of diffusion prior model~\cite{rombach2022stablediffusion_cvpr}. Ideally, we want to specify the target individual observed from our input images, to encourage the diffusion model to synthesize the consistent human appearance at the virtual viewpoints.  
To incorporate this idea, we leverage the Textual Inversion (TI) by finding the text-embedding specific to our target human~\cite{gal2022textual_inversion, ruiz2023dreambooth}.
Here, we use CustomDiffusion~\cite{kumari2022customdiffusion} to invert observations of the target individual into the text token \texttt{<person-j>}, 
where we collect the images of the target human from input frames by cropping and masking the target person only. 
Together with Textual Inversion, CustomDiffusion fine-tunes the attention and text embedding layer of the diffusion $\phi$ which makes a person-specific diffusion $\phi_j$.
By adding the inverted text-token to the text-prompt, we can get diffusion-generated images consistent with the observations.
We perform the diffusion fine-tuning and Textual Inversion per person separately and apply the subject-specific fine-tuned version of SDS for each target person.

Furthermore, we also utilize the OpenPose ControlNet~\cite{zhang2023controlnet} to align the body pose of a person generated by diffusion and our person Gaussians $\mathcal{G}^h$. For this, when we compute the SDS loss, we project the 3D SMPL joints $J_{smpl}(\vec{\theta},\vec{\beta}_j)$ into the viewpoint, convert to OpenPose~\cite{Cao2019openpose} format, and query them into the ControlNet.
 We additionally add a view-augmented language prompt~\cite{poole2022dreamfusion} for stable optimization. 
 We sample the noise time-step $\tau$ from $\mathcal{U}[0.2, 0.98]$ for the first 2000 iterations and then we decay the maximum time-step from $0.98$ to $0.3$ for 2000 iterations.
 Also to enhance the fine details of the body, we randomly sample camera $v_j$ which zooms in the face, lower body, and upper body. 
 Refer to our supp. mat. for more details.

\subsection{Training Objectives}
 For every iteration, we render the image $R_t$ corresponding to frame $t$ and calculate MSE loss, SSIM loss, and LPIPS loss by comparing it with the ground truth image $I_t$: 
\begin{multline}
    \mathcal{L}_{recon} = \lambda_{rgb}\text{MSE}(R_t, I_t) + \lambda_{ssim}\text{SSIM}(R_t, I_t) \\
    + \lambda_{lpips}\text{LPIPS}(R_t, I_t).
\end{multline}
 Then we compute SDS Loss $\mathcal{L}_{SDS}$ for each person's Gaussians $\mathcal{G}^h_j$ with person-specific diffusion $\phi_j$:
\begin{equation}
    \mathcal{L}_{sds} = \sum_{j=1}^{N} \mathcal{L}_{SDS}(\mathcal{G}^h_j, \phi_j).
\end{equation}
 To avoid transparent artifacts, we additionally add hard-surface regularization on human rendering following LOLNeRF\cite{rebain2022lolnerf}:
 \begin{equation}
    \mathcal{L}_{hard} = -\log (\exp^{-|\alpha|} + \exp^{|\alpha|}) + \text{const.}
 \end{equation},
 where $\alpha$ is the rendered alpha map of person Gaussian $\mathcal{G}^h_j$. Our final training objective is as follows: 
 \begin{equation}
    \mathcal{L}_{tot} = \lambda_{recon}\mathcal{L}_{recon} +  \lambda_{sds}\mathcal{L}_{sds} + \lambda_{hard}\mathcal{L}_{hard}.
 \end{equation}

\begin{figure*}[ht!]
    \centering
    \includegraphics[width=1.\linewidth, trim={0 5.5cm 0 0.5cm},clip]{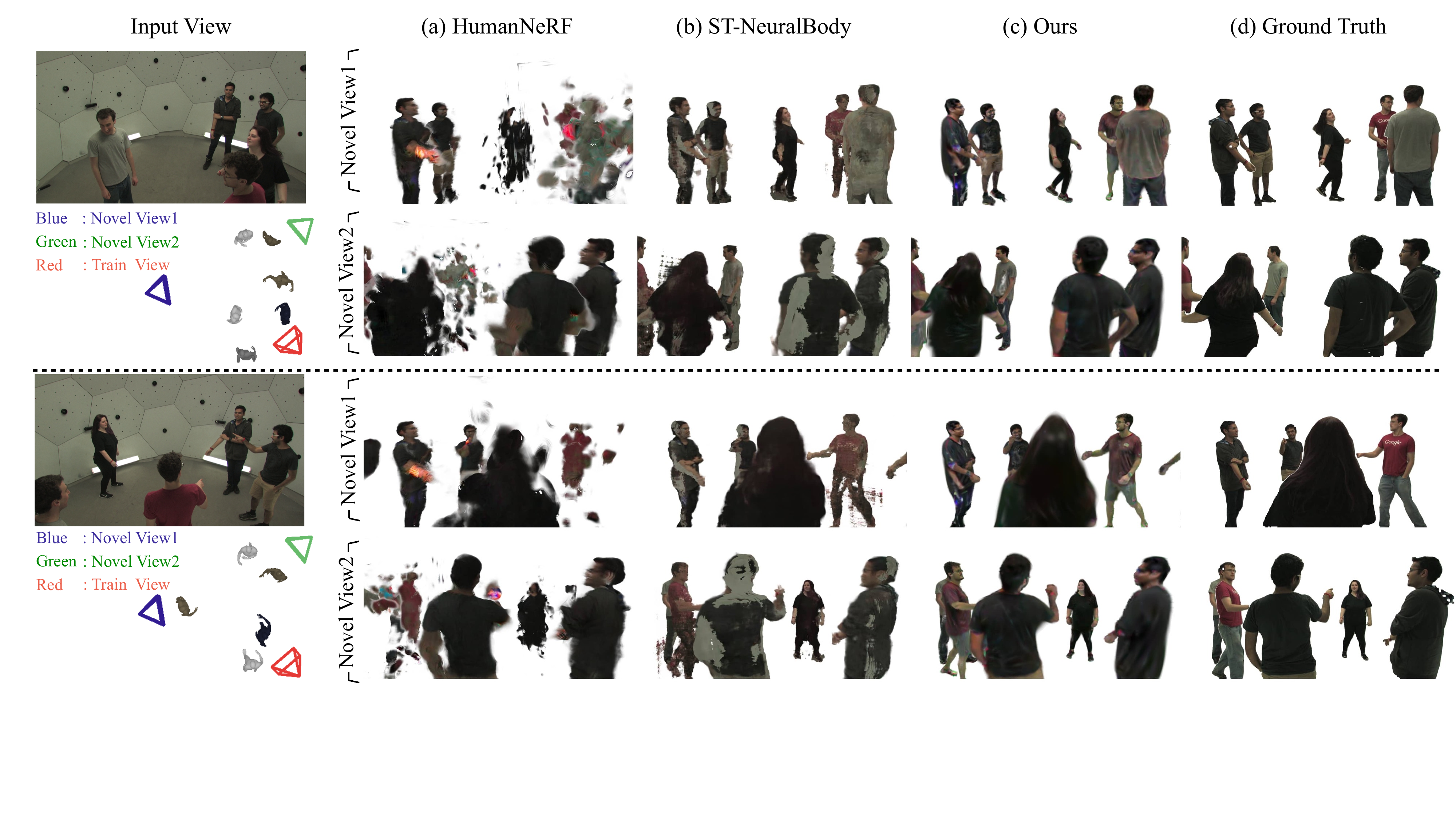}
    \vspace{-20px}
    \caption{\textbf{Novel view synthesized results of Panoptic Dataset~\cite{panoptic_tpami}}}
    \label{fig:panoptic_result}
\vspace{-10pt}
\end{figure*}

\section{Experiments}
\label{sec:experiments}
We perform rigorous quantitative and qualitative evaluations to show the strengths of our method. We first test our method on challenging scenes capturing multiple people with sparse 2D observations, to show the major advantage of our method. We also apply our method to existing single human reconstruction benchmarks. Additionally, by performing ablation studies, we demonstrate the efficacy of each module within our pipeline. We also show the computational efficiency of our method by comparing the rendering time to the existing method.

\subsection{Dataset}
\noindent{\textbf{Panoptic Dataset~\cite{panoptic_tpami}}} This dataset captures socially interacting multiple individuals via a multi-view camera system. We simulate the sparse 2D view scenario by choosing a single camera view as the input for the reconstruction and using other views for GT views. 
We use an ultimatum sequence with 6 people (540 frames in ultimatum 160422 sequence), and choose an HD camera view in a common view angle as the input video.
We select other 7 HD cameras in diverse novel viewpoints as the GTs for the evaluations, as shown in \cref{fig:panoptic_result}.
As a pre-processing, we fit SMPL models on the provided pseudo-GT 3D skeletons of the dataset using a pose-prior~\cite{bogo2016smplify}.
See the supp. mat. for more details of processing.

\noindent{\textbf{Hi4D~\cite{yin2023hi4d}}} This dataset contains multiple individuals at close distances, which are captured with synchronized 8 cameras. 
We consider challenging two sequences \texttt{pair00-dance} and \texttt{pair01-hug}.
Similar to Panoptic DB, we choose the video from a camera (camera 76) as input where only a single side of the people is visible and use all other 7 views as novel GT views for the evaluation. We use the provided pseudo-GT SMPL parameters. 

\noindent{\textbf{Single Human Benchmarks}} 
We also conduct experiments on an existing single human benchmark ZJU-Mocap~\cite{peng2021neuralbody_zjumocap} dataset, to check the performance of our model on the single human reconstruction task with sufficient observations. We follow the evaluation pipeline used in baselines~\cite{Weng2022humannerf,jiang2023instantavatar}.

\subsection{Baseline and Evaluation Metrics}
 We compare our model with three methods, \textbf{HumanNeRF}~\cite{Weng2022humannerf}, \textbf{InstantAvatar}~\cite{jiang2023instantavatar} and \textbf{Shuai et al.}~\cite{Shuai2022soccer_dataset}.
\textbf{HumanNeRF}~\cite{Weng2022humannerf} shows SOTA quality on the monocular human reconstruction task.
As HumanNeRF cannot handle multiple people at once, we optimize it separately for each person and merge them in the final evaluation. We get a rendering of each individual separately and accumulate them in an $\alpha$ blending manner. The order of accumulation is determined by the distance of the SMPL pelvis from the center of the camera. 
Because HumanNeRF requires a foreground mask for processing, we use GT masks if available (Hi4D and ZJU-Mocap), or use an off-the-shelf method~\cite{kirillov2023segment}. 
\textbf{InstantAvatar}~\cite{jiang2023instantavatar} is the most efficient method to reconstruct a human from a monocular video with high quality. We evaluate the InstantAvatar using the same pipeline applied for HumanNeRF evaluation, as it's only capable of mono-human cases. 
\textbf{Shuai et al.}~\cite{Shuai2022soccer_dataset} is a compositional method similar to ours, which optimizes implicit functions of people and background from sparse view input~\cite{Shuai2022soccer_dataset}. It represents each person with NeuralBody~\cite{peng2021neuralbody_zjumocap} and background with NeRF~\cite{mildenhall2020nerf} and renders them together by using the compositional rendering pipeline of ST-NeRF~\cite{zhang2021editable_stnerf}. Different from the original paper, we optimize it using only a single train view. 

 We compare the quality of human rendering by masking out the background of GT images. For evaluation metric, we use peak signal-to-noise ratio (PSNR), structural similarity (SSIM), and perceptual similarity (LPIPS)~\cite{zhang2018perceptual} following the prior work~\cite{Weng2022humannerf}.

\begin{table}[t]
    \small
    \centering
    \resizebox{\linewidth}{!}
    {
        \begin{tabular}{lccc}
    	\toprule
        Methods &PSNR$\uparrow$  &SSIM$\uparrow$ & LPIPS* $\downarrow$ \\
        \midrule
        HumanNeRF~\cite{Weng2022humannerf}   & 19.59 & 0.6514 & 38.69 \\
        InstantAvatar~\cite{jiang2023instantavatar} & 15.03 & 0.4163 & 65.95 \\
        \emph{Shuai et al.} ~\cite{Shuai2022soccer_dataset} & 15.79 & {\textbf{0.8370}} & 25.77 \\
        \midrule
        Ours & {\bf 23.60} & 0.8323 & \textbf{25.41 }\\
        \bottomrule
        \end{tabular} 
    }
    \vspace{-5pt}
    \caption{\textbf{Quantitative results on Panoptic dataset.} LPIPS* = $100\times\text{LPIPS}$. Our method shows better performance in PSNR and LPIPS and comparable results in SSIM. 
} 
\vspace{-5pt}
\label{tab:quant-panoptic}
\end{table}

\subsection{Implementation Details}
To stabilize optimization, we first train background Gaussians $\mathcal{G}^{BG}$ solely with background images where people are masked out. We then optimize human Gaussians $\mathcal{G}_{p_i,c}$ without $\mathcal{L}_{SDS}$ for the first 1000 iterations and then, optimize human Gaussians $\mathcal{G}^{p_i}_{c}$ and background Gaussians $\mathcal{G}_{bg}$ simultaneously together with $\mathcal{L}_{SDS}$ for 10k iterations. 
Refer to our supp. mat. for more details.

\begin{table}[t]
    \small
    \centering
    \resizebox{\linewidth}{!}{
        \begin{tabular}{lcccc}
        \toprule
	\multirow{2.5}{*}{Method}
    &\multicolumn{2}{c}{\texttt{pair00-dance}}&\multicolumn{2}{c}{\texttt{pair01-hug}} \\
\cmidrule(lr){2-3}\cmidrule(lr){4-5} &PSNR$\uparrow$ &SSIM$\uparrow$&PSNR$\uparrow$ &SSIM$\uparrow$ \\
        \midrule
        HumanNeRF~\cite{Weng2022humannerf}   & 18.79 & 0.8552 & 21.40 & 0.8238 \\
        InstantAvatar~\cite{jiang2023instantavatar}   & 18.60 & 0.8646 & 19.07 & 0.8254 \\
        \emph{{Shuai et al.}~\cite{Shuai2022soccer_dataset}} & 20.78 & 0.9165 & 19.72 & 0.9078 \\
        \midrule
        Ours  & {\bf 23.76} & {\bf 0.9328}  & {\bf 25.14}  & {\bf 0.9289} \\
        \bottomrule
        \end{tabular}
}
\vspace{-5pt}
\caption{\textbf{Quantitative results on Hi4D dataset~\cite{yin2023hi4d}}. Ours show better performance on both PSNR and SSIM in situations when people closely interact like hugging or dancing together}
\vspace{-15pt}
\label{tab:quant_hi4d}
\end{table}

\subsection{Evaluations on Multiple People Reconstruction}
\noindent\textbf{Panoptic dataset.} 
We show the qualitative comparison between our results and the outputs of baselines in \figref{fig:panoptic_result}. As shown, our method reconstructs the appearances of people even in the presence of severe occlusions and image cropping. In contrast, both HumanNeRF~\cite{Weng2022humannerf} and \emph{Shuai et al.}~\cite{Shuai2022soccer_dataset} fail to reconstruct details showing noticeable artifacts since many body parts are not visible in the input view.
In particular, HumanNeRF suffers from severe occlusions because it requires accurate foreground masks containing whole human shapes, which cannot be obtained due to the occluder in front of the target individual.  
In contrast, our method does not suffer from such issues. 

We also quantify the output qualities in \tabref{tab:quant-panoptic} by rendering the reconstructed scenes into unseen novel views. As shown in the table, our method outperforms the baselines in PSNR and LPIPS.

\begin{figure}[t]
     \centering
        \includegraphics[width=1.0\columnwidth, trim={0 0 0 0},clip]{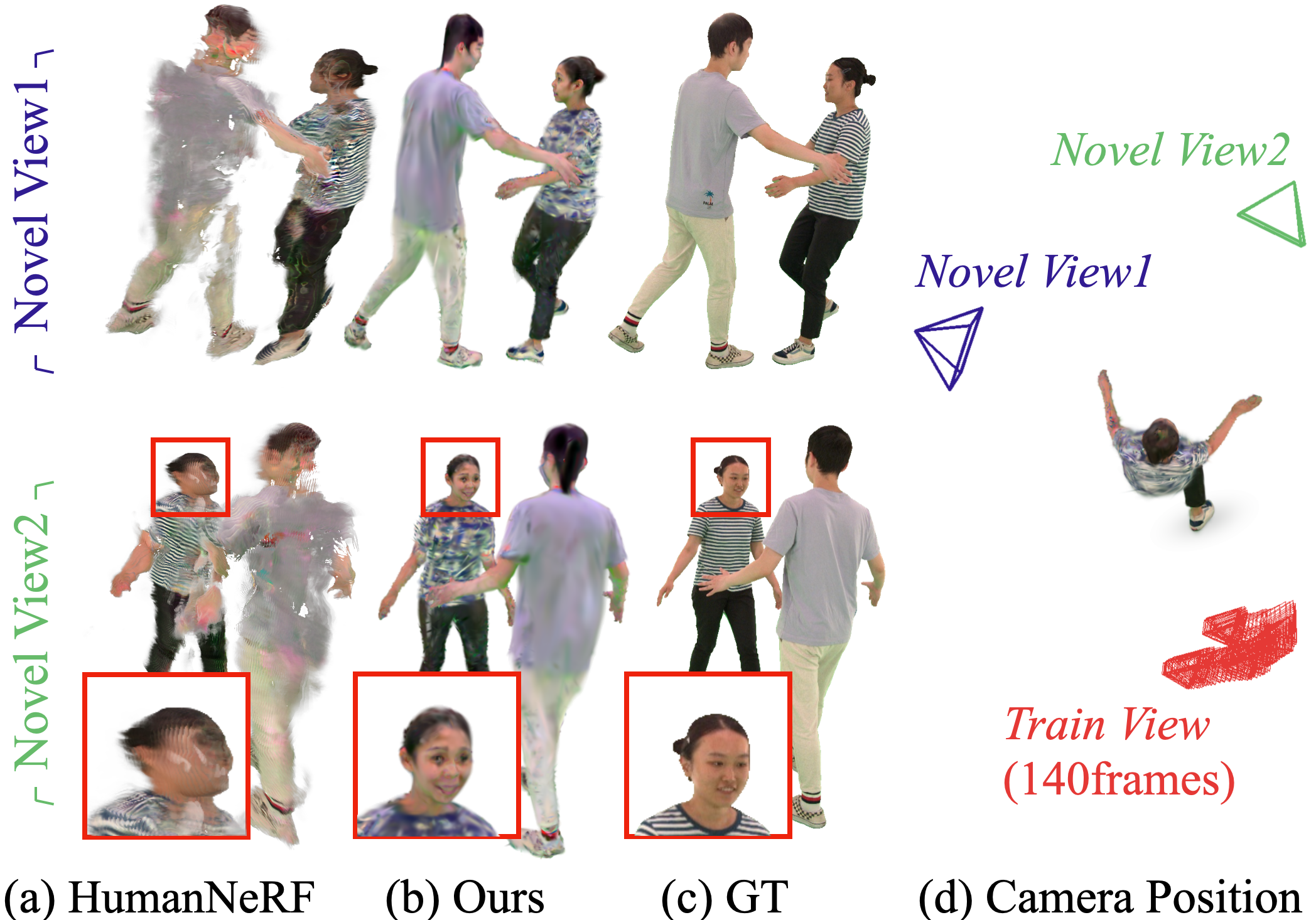}
     \hfill
     \vspace{-15pt}
    \caption{\textbf{Novel view synthesis output of Hi4D \textit{pair00-dance} sequence.} While HumanNeRF\cite{Weng2022humannerf} fails to reconstruct a face, ours synthesizes a plausible face guided by diffusion model~\cite{rombach2022stablediffusion_cvpr}. (d) plots camera position relative to the front viewing female body. As shown here, the majority of the rendered output shown here has been never observed in the train view. }
\vspace{-10pt}
\label{fig:hi4d_qualitative}
\end{figure}

\noindent\textbf{Hi4D dataset.}
We show the quantitative results in \tabref{tab:quant_hi4d} and qualitative examples in \figref{fig:hi4d_qualitative} from \texttt{pair00-dance} sequence.
As shown in \tabref{tab:quant_hi4d}, our method outperforms baselines in all metrics. 
Interestingly, in this dataset, some body parts of the individual are never observed in the input view due to the severe occlusions, e.g.,  the face of the female person in \figref{fig:hi4d_qualitative}, where our method ``hallucinates'' realistic human face without any observations. 

\begin{table}[t]
    \centering
    \resizebox{\linewidth}{!}
    {
        \small
        \begin{tabular}{lccc}
    	\toprule
        Method & PSNR$\uparrow$ &SSIM$\uparrow$  &LPIPS*$\downarrow$ \\
        \midrule
        HumanNeRF~\cite{Weng2022humannerf}  & 30.23 & 0.9554	& 3.36 \\
        InstantAvatar ~\cite{jiang2023instantavatar}  & 28.55 & 0.9282 & 10.60 \\
        \midrule
        Ours \textit{wo} $\mathcal{L}_{SDS}$  & 30.10 & 0.9529 & 4.68 \\
        Ours & 30.13 & 0.9535 & 4.50 \\
        \bottomrule
        \end{tabular}
}
\vspace{-5pt}
\caption{\textbf{Quantitative results on 6 subjects in ZJU-Mocap dataset~\cite{peng2021neuralbody_zjumocap}}. LPIPS* = $100\times\text{LPIPS}$. Our method shows comparable quality compared to HumanNeRF~\cite{Weng2022humannerf} even if the observation is sufficient.}
\vspace{-5pt}
\label{tab:quant-zjumocap}
\end{table}

\subsection{Evaluations on Single Person Reconstruction} 
We show the quantitative comparisons on ZJU-Mocap~\cite{peng2021neuralbody_zjumocap} dataset. 
As shown in \tabref{tab:quant-zjumocap}, our method shows comparable performance over baselines when sufficient 2D observations are available. This result demonstrates that our pipeline based on 3D-GS with SDS loss does not negatively affect the single-person reconstruction scenarios while showing its major strengths in the case of sparse observations.

\begin{figure}[t]
    \centering
    \includegraphics[width=1.0\columnwidth, trim={0 0 0 0},clip]{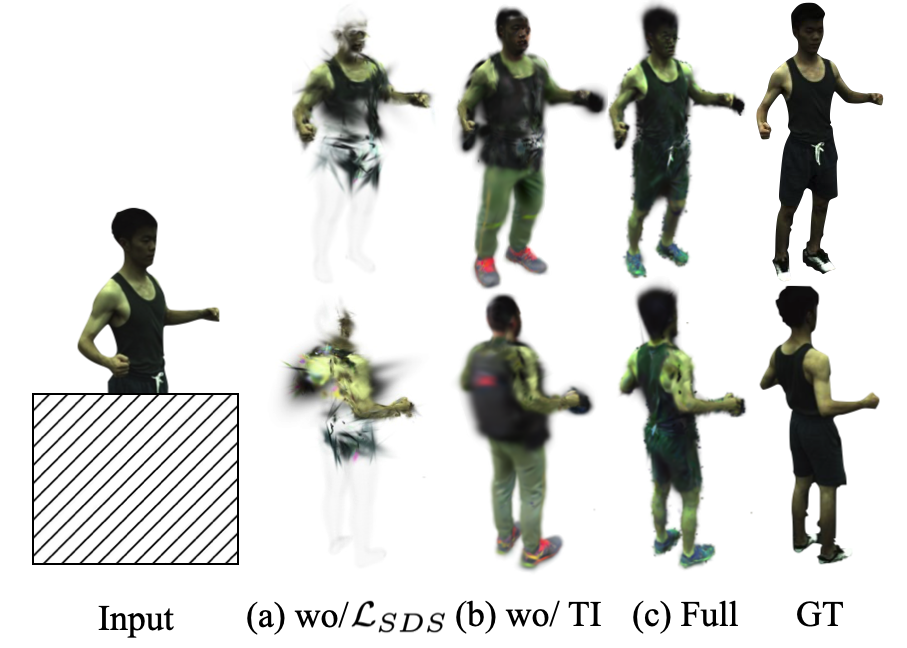}
    \vspace{-15pt}
    \caption{\textbf{Ablation studies.} From (a) shows that using only reconstruction loss suffers from artifacts (like hedgehogs) even in shown regions. (b) shows that textual inversion is essential to generate contextually similar appearances on unseen regions. }
    \label{fig:ablation}
\vspace{-5pt}
\end{figure}
\begin{table}[t]
    \small
    \centering
    \resizebox{\linewidth}{!}{
        \begin{tabular}{ccc|ccc}
    	\toprule
             w/ Textual Inversion & w/ $\mathcal{L}_{SDS}$  & w/ $\mathcal{L}_{recon}$  & PSNR$\uparrow$ & SSIM$\uparrow$ & LPIPS* $\downarrow$ \\
            \midrule
            \checkmark & \checkmark & \checkmark & \bf{26.03} & \bf{0.9435} & {\bf 7.00} \\
             & \checkmark & \checkmark & 23.43 & 0.9342 & 8.36 \\
             &   & \checkmark  & 24.51 & 0.9323 & 10.54 \\
        \bottomrule
        \end{tabular}
    }
\vspace{-5pt}
\caption{\textbf{Ablation Studies with lower-body occluded ZJU-Mocap~\cite{peng2021neuralbody_zjumocap} 377 subject.} LPIPS* = $100\times$LPIPS. As shown, SDS loss with textual inversion token shows the biggest improvement.} 
\label{tab:ablation}
\vspace{-15pt}
\end{table}

\subsection{Ablation Studies}
We conduct an ablation study to demonstrate the importance of the proposed modules of our framework. 
As a way for the quantitative evaluations, we use ZJU-Mocap~\cite{peng2021neuralbody_zjumocap} dataset and simulate a challenging scenario by (1) completely occluding the lower body, and (2) using a few frames only for the input (10\% of frames from the ZJU-Mocap 377 subject). An example of the artificially occluded images is shown in \figref{fig:ablation} with test results.

Our \emph{Full} method can successfully reconstruct the whole part of the humans. Interestingly, it also synthesizes the completely unseen short pants and shoes of the target individual, while the appearance and colors are different from the GT. 
As expected, the output without $\mathcal{L}_{SDS}$ fails to reconstruct the unseen lower part,  and also shows poor quality on the upper body due to the insufficient image frames.
The output without Texture Inversion (TI) reconstructs the unseen lower parts as well, but, interestingly, the output appearance is very different from the GT. This result directly demonstrates the importance of our Texture Inversion process in applying SDS loss, helpful in preserving the identity of the target individual. We also show the quantification results in \tabref{tab:ablation}, where the importance of each module is also clearly demonstrated.

\begin{table}[t]
    \small
    \centering
    \resizebox{\linewidth}{!}
    {
        \begin{tabular}{l|ccc}
    	\toprule
        \multirow{2.5}{*}{Method}&\multicolumn{2}{c}{Rendering Speed (FPS $\uparrow$)} & \multirow{2.5}{*}{VRAM $\downarrow$} \\
        \cmidrule(lr){2-3} & $512\times 512$ & $1024\times 1024$  \\
        \midrule
        InstantAvatar~\cite{jiang2023instantavatar}  & 18.03 & 7.09 & 4972MiB \\
        Ours (15k) & {\bf 361.01} & {\bf 277.01} & {\bf 1496MiB} \\
        \bottomrule
        \end{tabular}
}
\vspace{-5pt}
\caption{\textbf{Novel pose rendering speed.} We compared novel pose rendering speed between InstantAvatar~\cite{jiang2023instantavatar} and ours with ZJU-Mocap~\cite{peng2021neuralbody_zjumocap}
 377 subject. %
It shows that our method consists of a 15k Gaussians renders faster on both low and high resolutions, with less VRAM consumption.}
\vspace{-15pt}
\label{tab:rendering-speed}
\end{table}

\subsection{Rendering Efficiency}
We show the computational efficiency of our method by comparing the rendering time of novel pose synthesis to InstantAvatar~\cite{jiang2023instantavatar}, which is known as the most efficient existing method and also the fastest among our competitors.
All reported times here are measured with a single GeForce RTX 3090 GPU.
By taking advantage of 3D-GS representations, our method achieves more than real-time rendering speed with 300 FPS on 1K$\times$1K images.
As shown in \tabref{tab:rendering-speed}, our method surpasses InstantAvatar~\cite{jiang2023instantavatar}  both in rendering speed and memory consumption while showing better rendering quality as shown in \tabref{tab:quant-zjumocap}.

\section{Discussion}
\label{sec:conclusion}
In this paper, we present a method to reconstruct the world and dynamically moving humans in 3D from a monocular video input, particularly focusing on sparse and limited observation scenarios.
We represent both the world and multiple humans via 3D Gaussian Splatting representation, enabling us to conveniently and efficiently compose and render them together. We also introduce a novel approach to optimize the 3D-GS representation in a canonical space by fusing the sparse cues in the common space, where we leverage a pre-trained 2D diffusion model to synthesize unseen views by keeping the consistency with the observed 2D appearances. 
Via thorough experiments, we demonstrate the high performance and efficiency of our method in various challenging examples.

Our approach, however, still has limitations such as: (1) SMPL fitting needs to be provided; (2) our method only considers humans as the dynamic target, ignoring animals, cars, or other dynamic objects; (3) the quality of the synthesized parts are still limited with visible artifacts. All these limitations can be exciting future research directions.

\noindent \textbf{Acknowledgements} 
This work was supported by Samsung Electronics C-Lab, NRF grant funded by the Korea government (MSIT) (No. 2022R1A2C2092724 and No. RS-2023-00218601), and IITP grant funded by the Korean government (MSIT) (No.2021-0-01343). H. Joo is the corresponding author.

{\small
\bibliographystyle{ieee_fullname}
\bibliography{11_references}

\begin{thebibliography}{10}\itemsep=-1pt

\bibitem{Akhter2012BilinearBasis}
Ijaz Akhter, Tomas Simon, Sohaib Khan, Iain Matthews, and Yaser Sheikh.
\newblock Bilinear spatiotemporal basis models.
\newblock {\em TOG}, 2012.

\bibitem{Alldieck2018people_snaphost}
Thiemo Alldieck, Marcus Magnor, Weipeng Xu, Christian Theobalt, and Gerard Pons-Moll.
\newblock Video based reconstruction of 3d people models.
\newblock In {\em CVPR}, 2018.

\bibitem{bogo2016smplify}
Federica Bogo, Angjoo Kanazawa, Christoph Lassner, Peter Gehler, Javier Romero, and Michael~J. Black.
\newblock Keep it {SMPL}: Automatic estimation of {3D} human pose and shape from a single image.
\newblock In {\em ECCV}, 2016.

\bibitem{Cao2019openpose}
Z. {Cao}, G. {Hidalgo Martinez}, T. {Simon}, S. {Wei}, and Y.~A. {Sheikh}.
\newblock Openpose: Realtime multi-person 2d pose estimation using part affinity fields.
\newblock {\em TPAMI}, 2019.

\bibitem{Chan2023iccv_GeNVS}
Eric~R. Chan, Koki Nagano, Matthew~A. Chan, Alexander~W. Bergman, Jeong~Joon Park, Axel Levy, Miika Aittala, Shalini De~Mello, Tero Karras, and Gordon Wetzstein.
\newblock Generative novel view synthesis with 3d-aware diffusion models.
\newblock In {\em ICCV}, 2023.

\bibitem{Chen_2023iccv_fantasia3d}
Rui Chen, Yongwei Chen, Ningxin Jiao, and Kui Jia.
\newblock Fantasia3d: Disentangling geometry and appearance for high-quality text-to-3d content creation.
\newblock In {\em ICCV}, 2023.

\bibitem{chen2023fastsnarf}
Xu Chen, Tianjian Jiang, Jie Song, Max Rietmann, Andreas Geiger, Michael~J Black, and Otmar Hilliges.
\newblock Fast-snarf: A fast deformer for articulated neural fields.
\newblock {\em TPAMI}, 2023.

\bibitem{chen2021snarf}
Xu Chen, Yufeng Zheng, Michael~J Black, Otmar Hilliges, and Andreas Geiger.
\newblock Snarf: Differentiable forward skinning for animating non-rigid neural implicit shapes.
\newblock In {\em ICCV}, 2021.

\bibitem{deng2023nerdi}
Congyue Deng, Chiyu Jiang, Charles~R Qi, Xinchen Yan, Yin Zhou, Leonidas Guibas, Dragomir Anguelov, et~al.
\newblock Nerdi: Single-view nerf synthesis with language-guided diffusion as general image priors.
\newblock In {\em CVPR}, 2023.

\bibitem{gal2022textual_inversion}
Rinon Gal, Yuval Alaluf, Yuval Atzmon, Or Patashnik, Amit~H. Bermano, Gal Chechik, and Daniel Cohen-Or.
\newblock An image is worth one word: Personalizing text-to-image generation using textual inversion.
\newblock In {\em ICLR}, 2023.

\bibitem{goel2023humans_in_4d}
Shubham Goel, Georgios Pavlakos, Jathushan Rajasegaran, Angjoo Kanazawa*, and Jitendra Malik*.
\newblock Humans in 4{D}: Reconstructing and tracking humans with transformers.
\newblock In {\em ICCV}, 2023.

\bibitem{guo2023vid2avatar}
Chen Guo, Tianjian Jiang, Xu Chen, Jie Song, and Otmar Hilliges.
\newblock Vid2avatar: 3d avatar reconstruction from videos in the wild via self-supervised scene decomposition.
\newblock In {\em CVPR}, 2023.

\bibitem{hz_mvgeometry}
R.~I. Hartley and A. Zisserman.
\newblock {\em Multiple View Geometry in Computer Vision}.
\newblock Cambridge University Press, ISBN: 0521540518, second edition, 2004.

\bibitem{ho2020ddpm}
Jonathan Ho, Ajay Jain, and Pieter Abbeel.
\newblock Denoising diffusion probabilistic models.
\newblock In {\em NeurIPS}, 2020.

\bibitem{huang2024tech}
Yangyi Huang, Hongwei Yi, Yuliang Xiu, Tingting Liao, Jiaxiang Tang, Deng Cai, and Justus Thies.
\newblock {TeCH: Text-guided Reconstruction of Lifelike Clothed Humans}.
\newblock In {\em 3DV}, 2024.

\bibitem{jiang2022selfrecon}
Boyi Jiang, Yang Hong, Hujun Bao, and Juyong Zhang.
\newblock Selfrecon: Self reconstruction your digital avatar from monocular video.
\newblock In {\em CVPR}, 2022.

\bibitem{jiang2023instantavatar}
Tianjian Jiang, Xu Chen, Jie Song, and Otmar Hilliges.
\newblock Instantavatar: Learning avatars from monocular video in 60 seconds.
\newblock In {\em CVPR}, 2023.

\bibitem{jiang2022neuman}
Wei Jiang, Kwang~Moo Yi, Golnoosh Samei, Oncel Tuzel, and Anurag Ranjan.
\newblock Neuman: Neural human radiance field from a single video.
\newblock In {\em ECCV}, 2022.

\bibitem{panoptic_tpami}
Hanbyul Joo, Tomas Simon, Xulong Li, Hao Liu, Lei Tan, Lin Gui, Sean Banerjee, Timothy~Scott Godisart, Bart Nabbe, Iain Matthews, Takeo Kanade, Shohei Nobuhara, and Yaser Sheikh.
\newblock Panoptic studio: A massively multiview system for social interaction capture.
\newblock {\em TPAMI}, 2017.

\bibitem{kerbl20233Dgaussians}
Bernhard Kerbl, Georgios Kopanas, Thomas Leimk{\"u}hler, and George Drettakis.
\newblock 3d gaussian splatting for real-time radiance field rendering.
\newblock In {\em SIGGRAPH}, 2023.

\bibitem{kingma2014adam}
Diederik~P Kingma and Jimmy Ba.
\newblock Adam: A method for stochastic optimization.
\newblock In {\em ICLR}, 2015.

\bibitem{kirillov2023segment}
Alexander Kirillov, Eric Mintun, Nikhila Ravi, Hanzi Mao, Chloe Rolland, Laura Gustafson, Tete Xiao, Spencer Whitehead, Alexander~C Berg, Wan-Yen Lo, et~al.
\newblock Segment anything.
\newblock In {\em ICCV}, 2023.

\bibitem{kumari2022customdiffusion}
Nupur Kumari, Bingliang Zhang, Richard Zhang, Eli Shechtman, and Jun-Yan Zhu.
\newblock Multi-concept customization of text-to-image diffusion.
\newblock In {\em CVPR}, 2023.

\bibitem{Abhijit2022PNF}
Abhijit Kundu, Kyle Genova, Xiaoqi Yin, Alireza Fathi, Caroline Pantofaru, Leonidas Guibas, Andrea Tagliasacchi, Frank Dellaert, and Thomas Funkhouser.
\newblock {Panoptic Neural Fields: A Semantic Object-Aware Neural Scene Representation}.
\newblock In {\em CVPR}, 2022.

\bibitem{lei2023rgbd2}
Jiabao Lei, Jiapeng Tang, and Kui Jia.
\newblock Rgbd2: Generative scene synthesis via incremental view inpainting using rgbd diffusion models.
\newblock In {\em CVPR}, 2023.

\bibitem{liao2023tada_3dv_accepted}
Tingting Liao, Hongwei Yi, Yuliang Xiu, Jiaxaing Tang, Yangyi Huang, Justus Thies, and Michael~J Black.
\newblock Tada! text to animatable digital avatars.
\newblock In {\em 3DV}, 2024.

\bibitem{lin2023magic3d}
Chen-Hsuan Lin, Jun Gao, Luming Tang, Towaki Takikawa, Xiaohui Zeng, Xun Huang, Karsten Kreis, Sanja Fidler, Ming-Yu Liu, and Tsung-Yi Lin.
\newblock Magic3d: High-resolution text-to-3d content creation.
\newblock In {\em CVPR}, 2023.

\bibitem{liu2023zero123}
Ruoshi Liu, Rundi Wu, Basile Van~Hoorick, Pavel Tokmakov, Sergey Zakharov, and Carl Vondrick.
\newblock Zero-1-to-3: Zero-shot one image to 3d object.
\newblock In {\em ICCV}, 2023.

\bibitem{smpl2015}
Matthew Loper, Naureen Mahmood, Javier Romero, Gerard Pons-Moll, and Michael~J. Black.
\newblock Smpl: A skinned multi-person linear model.
\newblock {\em TOG}, 2015.

\bibitem{lugmayr2022repaint}
Andreas Lugmayr, Martin Danelljan, Andres Romero, Fisher Yu, Radu Timofte, and Luc Van~Gool.
\newblock Repaint: Inpainting using denoising diffusion probabilistic models.
\newblock In {\em CVPR}, 2022.

\bibitem{mahmood2019amass}
Naureen Mahmood, Nima Ghorbani, Nikolaus~F. Troje, Gerard Pons-Moll, and Michael~J. Black.
\newblock {AMASS}: Archive of motion capture as surface shapes.
\newblock In {\em ICCV}, 2019.

\bibitem{melas2023realfusion}
Luke Melas-Kyriazi, Iro Laina, Christian Rupprecht, and Andrea Vedaldi.
\newblock Realfusion: 360deg reconstruction of any object from a single image.
\newblock In {\em CVPR}, 2023.

\bibitem{mildenhall2020nerf}
Ben Mildenhall, Pratul~P. Srinivasan, Matthew Tancik, Jonathan~T. Barron, Ravi Ramamoorthi, and Ren Ng.
\newblock Nerf: Representing scenes as neural radiance fields for view synthesis.
\newblock In {\em ECCV}, 2020.

\bibitem{mueller2022instantngp}
Thomas M\"uller, Alex Evans, Christoph Schied, and Alexander Keller.
\newblock Instant neural graphics primitives with a multiresolution hash encoding.
\newblock In {\em SIGGRAPH}, 2022.

\bibitem{Neverova2020cse}
Natalia Neverova, David Novotny, Vasil Khalidov, Marc Szafraniec, Patrick Labatut, and Andrea Vedaldi.
\newblock Continuous surface embeddings.
\newblock In {\em NeurIPS}, 2020.

\bibitem{Julian2021NSG}
Julian Ost, Fahim Mannan, Nils Thuerey, Julian Knodt, and Felix Heide.
\newblock Neural scene graphs for dynamic scenes.
\newblock In {\em CVPR}, 2021.

\bibitem{hspark2010nrsfm}
Hyun~Soo Park, Takaaki Shiratori, Iain Matthews, and Yaser Sheikh.
\newblock 3d reconstruction of a moving point from a series of 2d projections.
\newblock In {\em ECCV}, 2010.

\bibitem{peng2021neuralbody_zjumocap}
Sida Peng, Yuanqing Zhang, Yinghao Xu, Qianqian Wang, Qing Shuai, Hujun Bao, and Xiaowei Zhou.
\newblock Neural body: Implicit neural representations with structured latent codes for novel view synthesis of dynamic humans.
\newblock In {\em CVPR}, 2021.

\bibitem{poole2022dreamfusion}
Ben Poole, Ajay Jain, Jonathan~T. Barron, and Ben Mildenhall.
\newblock Dreamfusion: Text-to-3d using 2d diffusion.
\newblock In {\em ICLR}, 2023.

\bibitem{rajasegaran2022tracking_phalp}
Jathushan Rajasegaran, Georgios Pavlakos, Angjoo Kanazawa, and Jitendra Malik.
\newblock Tracking people by predicting 3{D} appearance, location \& pose.
\newblock In {\em CVPR}, 2022.

\bibitem{rebain2022lolnerf}
Daniel Rebain, Mark Matthews, Kwang~Moo Yi, Dmitry Lagun, and Andrea Tagliasacchi.
\newblock Lolnerf: Learn from one look.
\newblock In {\em CVPR}, 2022.

\bibitem{rombach2022stablediffusion_cvpr}
Robin Rombach, Andreas Blattmann, Dominik Lorenz, Patrick Esser, and Bj{\"o}rn Ommer.
\newblock High-resolution image synthesis with latent diffusion models.
\newblock In {\em CVPR}, 2022.

\bibitem{rombach2022LDM}
Robin Rombach, Andreas Blattmann, Dominik Lorenz, Patrick Esser, and Bj{\"o}rn Ommer.
\newblock High-resolution image synthesis with latent diffusion models.
\newblock In {\em CVPR}, 2022.

\bibitem{ruiz2023dreambooth}
Nataniel Ruiz, Yuanzhen Li, Varun Jampani, Yael Pritch, Michael Rubinstein, and Kfir Aberman.
\newblock Dreambooth: Fine tuning text-to-image diffusion models for subject-driven generation.
\newblock In {\em CVPR}, 2023.

\bibitem{saharia2022photorealistic}
Chitwan Saharia, William Chan, Saurabh Saxena, Lala Li, Jay Whang, Emily~L Denton, Kamyar Ghasemipour, Raphael Gontijo~Lopes, Burcu Karagol~Ayan, Tim Salimans, et~al.
\newblock Photorealistic text-to-image diffusion models with deep language understanding.
\newblock In {\em NeurIPS}, 2022.

\bibitem{schoenberger2016sfm_colmap}
Johannes~Lutz Sch\"{o}nberger and Jan-Michael Frahm.
\newblock Structure-from-motion revisited.
\newblock In {\em CVPR}, 2016.

\bibitem{shao2022diffustereo}
Ruizhi Shao, Zerong Zheng, Hongwen Zhang, Jingxiang Sun, and Yebin Liu.
\newblock Diffustereo: High quality human reconstruction via diffusion-based stereo using sparse cameras.
\newblock In {\em ECCV}, 2022.

\bibitem{Shuai2022soccer_dataset}
Qing Shuai, Chen Geng, Qi Fang, Sida Peng, Wenhao Shen, Xiaowei Zhou, and Hujun Bao.
\newblock Novel view synthesis of human interactions from sparse multi-view videos.
\newblock In {\em SIGGRAPH}, 2022.

\bibitem{sun2021romp}
Yu Sun, Qian Bao, Wu Liu, Yili Fu, Michael~J Black, and Tao Mei.
\newblock Monocular, one-stage, regression of multiple 3d people.
\newblock In {\em ICCV}, 2021.

\bibitem{teed2021droidslam}
Zachary Teed and Jia Deng.
\newblock {DROID-SLAM: Deep Visual SLAM for Monocular, Stereo, and RGB-D Cameras}.
\newblock In {\em NeurIPS}, 2021.

\bibitem{wang2023sjc}
Haochen Wang, Xiaodan Du, Jiahao Li, Raymond~A Yeh, and Greg Shakhnarovich.
\newblock Score jacobian chaining: Lifting pretrained 2d diffusion models for 3d generation.
\newblock In {\em CVPR}, 2023.

\bibitem{wang2021neus}
Peng Wang, Lingjie Liu, Yuan Liu, Christian Theobalt, Taku Komura, and Wenping Wang.
\newblock Neus: Learning neural implicit surfaces by volume rendering for multi-view reconstruction.
\newblock In {\em NeurIPS}, 2021.

\bibitem{wang2023prolificdreamer}
Zhengyi Wang, Cheng Lu, Yikai Wang, Fan Bao, Chongxuan Li, Hang Su, and Jun Zhu.
\newblock Prolificdreamer: High-fidelity and diverse text-to-3d generation with variational score distillation.
\newblock In {\em NeurIPS}, 2023.

\bibitem{Weng2022humannerf}
Chung-Yi Weng, Brian Curless, Pratul~P. Srinivasan, Jonathan~T. Barron, and Ira Kemelmacher-Shlizerman.
\newblock Humannerf: Free-viewpoint rendering of moving people from monocular video.
\newblock In {\em CVPR}, 2022.

\bibitem{wu2022d2nerf}
Tianhao Wu, Fangcheng Zhong, Andrea Tagliasacchi, Forrester Cole, and Cengiz Oztireli.
\newblock D\^{} 2nerf: Self-supervised decoupling of dynamic and static objects from a monocular video.
\newblock In {\em NeurIPS}, 2022.

\bibitem{xiang2023occnerf}
Tiange Xiang, Adam Sun, Jiajun Wu, Ehsan Adeli, and Li Fei-Fei.
\newblock Rendering humans from object-occluded monocular videos.
\newblock In {\em ICCV}, 2023.

\bibitem{Hongyi2020GHUM}
Hongyi Xu, Eduard~Gabriel Bazavan, Andrei Zanfir, William~T. Freeman, Rahul Sukthankar, and Cristian Sminchisescu.
\newblock Ghum \& ghuml: Generative 3d human shape and articulated pose models.
\newblock In {\em CVPR}, 2020.

\bibitem{xu2018MonoPerfCap}
Weipeng Xu, Avishek Chatterjee, Michael Zollh\"{o}fer, Helge Rhodin, Dushyant Mehta, Hans-Peter Seidel, and Christian Theobalt.
\newblock Monoperfcap: Human performance capture from monocular video.
\newblock In {\em SIGGRAPH}, 2018.

\bibitem{xu2022vitpose}
Yufei Xu, Jing Zhang, Qiming Zhang, and Dacheng Tao.
\newblock Vi{TP}ose: Simple vision transformer baselines for human pose estimation.
\newblock In {\em NeurIPS}, 2022.

\bibitem{yang2022banmo}
Gengshan Yang, Minh Vo, Natalia Neverova, Deva Ramanan, Andrea Vedaldi, and Hanbyul Joo.
\newblock Banmo: Building animatable 3d neural models from many casual videos.
\newblock In {\em CVPR}, 2022.

\bibitem{yin2023hi4d}
Yifei Yin, Chen Guo, Manuel Kaufmann, Juan Zarate, Jie Song, and Otmar Hilliges.
\newblock Hi4d: 4d instance segmentation of close human interaction.
\newblock In {\em CVPR}, 2023.

\bibitem{yu2023monohuman}
Zhengming Yu, Wei Cheng, xian Liu, Wayne Wu, and Kwan-Yee Lin.
\newblock {MonoHuman}: Animatable human neural field from monocular video.
\newblock In {\em CVPR}, 2023.

\bibitem{zhang2021editable_stnerf}
Jiakai Zhang, Xinhang Liu, Xinyi Ye, Fuqiang Zhao, Yanshun Zhang, Minye Wu, Yingliang Zhang, Lan Xu, and Jingyi Yu.
\newblock Editable free-viewpoint video using a layered neural representation.
\newblock In {\em SIGGRAPH}, 2021.

\bibitem{zhang2023controlnet}
Lvmin Zhang, Anyi Rao, and Maneesh Agrawala.
\newblock Adding conditional control to text-to-image diffusion models.
\newblock In {\em ICCV}, 2023.

\bibitem{zhang2018perceptual}
Richard Zhang, Phillip Isola, Alexei~A Efros, Eli Shechtman, and Oliver Wang.
\newblock The unreasonable effectiveness of deep features as a perceptual metric.
\newblock In {\em CVPR}, 2018.

\bibitem{zhou2023sparsefusion}
Zhizhuo Zhou and Shubham Tulsiani.
\newblock Sparsefusion: Distilling view-conditioned diffusion for 3d reconstruction.
\newblock In {\em CVPR}, 2023.

\bibitem{zhu2023hifa}
Joseph Zhu and Peiye Zhuang.
\newblock Hifa: High-fidelity text-to-3d with advanced diffusion guidance.
\newblock In {\em ICLR}, 2024.

\bibitem{3DGS_gaussian_projection_theory}
M. Zwicker, H. Pfister, J. van Baar, and M. Gross.
\newblock Ewa volume splatting.
\newblock In {\em Proceedings Visualization, 2001. VIS '01.}, 2001.

\end{thebibliography}
}

\ifarxiv \clearpage \appendix

\section{Implementation Details}
\subsection{Baseline Implementation Details}
\noindent\textbf{HumanNeRF~\cite{Weng2022humannerf}} does not support the simultaneous optimization of multiple people, so we optimize each person separately and merge them in the evaluation stage. Following the default HumanNeRF experiment settings, each person is optimized for 400k iterations using 4 NVIDIA RTX4090 GPUs which takes approximately 40 hours per person. For the ZJU-Mocap~\cite{peng2021neuralbody_zjumocap} dataset, we utilize the publicly available checkpoints shared by the authors. 

\noindent\textbf{\emph{Shuai et al.}~\cite{Shuai2022soccer_dataset}} represents the scene as a composition of a background model and human model, both represented by a variant of NeRF~\cite{mildenhall2020nerf, peng2021neuralbody_zjumocap}. For the Panoptic dataset~\cite{panoptic_tpami} and Hi4D dataset~\cite{yin2023hi4d}, we model the background using a time-conditioned NeRF defined on the surface of the cylinder fully covering the scene and the human model with NeuralBody~\cite{peng2021neuralbody_zjumocap}. We jointly optimize these models for 400k iterations using 2 NVIDIA RTX4090 GPUs which takes approximately 70 hours per scene. The remaining settings are the same as the original paper~\cite{Shuai2022soccer_dataset}. When we render the scene for evaluation, we discard the background and only render the human model. 

\noindent\textbf{InstantAvatar~\cite{jiang2023instantavatar}} reconstructs a single person from monocular video input. Hence, we optimize it on each person separately and merge them in the evaluation stage same as HumanNeRF~\cite{Weng2022humannerf}. We train the InstantAvatar for 50 epochs using a single RTX3090, following the default options used to optimize PeopleSnapShot~\cite{Alldieck2018people_snaphost} in the original paper.

\subsection{Ours Implementation Details}
 \noindent\textbf{Background pre-optimization.} We first optimize background Gaussians $\mathcal{G}^{BG}$ with images that humans are masked out. The background Gaussians $\mathcal{G}^{BG}$ are initialized with point cloud obtained by SfM~\cite{schoenberger2016sfm_colmap} or SLAM~\cite{teed2021droidslam}. In the case of a fixed camera, we initialize Gaussians $\mathcal{G}^{BG}$ with a 3D sphere whose radius is 30m, together with background regularization loss to prevent it from occluding the people as follows:
 \begin{equation}
     \mathcal{L}_{reg}^{BG} = \lambda_{reg}^{BG} \sum_{i=0}^{N} ||\vec{\mu}_i^{BG} - 30||^2
 \end{equation},
 where $\vec{\mu}_i^{BG}$ is the center of $i$-th background Gaussian. We scale the world's unit distance to be 1m before starting optimization. The background is optimized for 30k iterations following the default 3D-GS~\cite{kerbl20233Dgaussians} experiment settings.
 
\noindent\textbf{Human background joint optimization.} 
After the pre-optimization of the background, we optimize human Gaussians ${\mathcal{G}^h_j}_{j=1,\dots, N}$ and background Gaussians $\mathcal{G}^{BG}$ together. For the first $1.5k$ iteration of joint optimization, we fix the center of human Gaussians $\vec{\mu}_i$ on the initial points $\vec{x}_{i,init}$ and clamp the opacity $o_i$ below $0.9$ to avoid the body being transparent. We densify the human Gaussians in $[2000, 2500, 3000]$ iterations for detailed reconstruction and prune Gaussians which are exceptionally large or transparent every 500 iterations until the end of optimizations to reduce artifacts. The background Gaussians are densified only during pre-optimization stage and keep the same number of Gaussians in the joint optimization stage. 

\noindent\textbf{Optimization Details.} 
We use Adam~\cite{kingma2014adam} optimizer with different learning rates for each component of 3D Gaussians. For the center of Gaussian $\vec{\mu}$, we set an initial learning rate as $1e^{-3}$ and decay it until $2e^{-6}$ during training. We use a fixed learning rate $2.5e^{-3}$ for color $\vec{c}$, $5e^{-2}$ for opacity $\vec{o}$, $5e^{-3}$ for scale $\vec{s}$, and $1e^{-3}$ for quaternion $\vec{q}$. 
We set the loss weight of SSIM loss $\lambda_{ssim}=0.2$, MSE loss $\lambda_{rgb}=0.8$, LPIPS loss $\lambda_{lpips}=0.1$, and SDS loss $\lambda_{sds}=1.0$. 
For hard surface regularization loss, we set the weight of loss $\lambda_{hard}$ relative to reconstruction loss weight $\lambda_{recon} = 0.1 \times \lambda_{recon}$ to keep a balance of losses.
We use a fixed reconstruction loss weight $\lambda_{recon}=1.0$ before $1k$ iterations and then schedule the weight after $1k$ iterations to balance the reconstruction loss and SDS loss.

\noindent\textbf{SDS loss details.} We use a publicly available SD1.5~\cite{rombach2022stablediffusion_cvpr} and OpenPose ControlNet~\cite{zhang2023controlnet} checkpoint for the SDS loss. Similar to other methods using SDS~\cite{poole2022dreamfusion}, we use a high CFG scale of 50 to generate detailed texture on unseen parts. We sample the noise time step $\tau$ of SDS loss from $\mathcal{U}[0.5, 0.98]$ for the first $2k$ iterations and then smoothly anneal it into $\mathcal{U}[0.02, 0.3]$ over following $2k$ iterations similar to the prior work~\cite{zhu2023hifa}. We also schedule the weight of reconstruction loss $\lambda_{recon}$ with a maximum time step $\tau_{max}$ on each iteration to balance the reconstruction loss and SDS loss as follows:
\begin{equation}
    \lambda_{recon} = 10^6 \times \tau_{max}^2.
\end{equation}
We apply SDS loss from $1k$ iteration of the joint optimization. For every single iteration of reconstruction loss, we apply SDS loss on all humans who appeared in the scene. 

We sample random unseen cameras for SDS loss from the surface of a sphere with a radius of 2.2, centered on the human pelvis. 
The azimuth $\varphi$ and elevation $\vartheta$ of cameras are drawn from $\varphi\sim \mathcal{U}[-\pi, \pi]$ and $\vartheta \sim \mathcal{U}[-0.3\pi, 0.3\pi]$. 
Additionally, we choose a view-augmented prompt [\textit{side, front, back}] based on the sampled azimuth $\varphi$ and SMPL global rotation. 
For the initial $3k$ iterations of optimization with SDS loss, we mainly render the full body of posed human Gaussians $\mathcal{G}^h_j(\vec{\theta}_{j,t})$ and canonical human Gaussians $\mathcal{G}^h_j(\vec{\theta}_{c})$ for SDS loss. In the subsequent iterations, we also randomly sample from zoomed-in views of the head, upper body, and lower body together with the full body of the posed human, and the full body of the canonical with a uniform probability of $0.2$. This two-stage random camera sampling facilitates the detailed reconstruction of unseen parts and head. 

\begin{figure}[t!]
    \centering
     \includegraphics[width=1.0\columnwidth, trim={0 0 -2 0},clip]{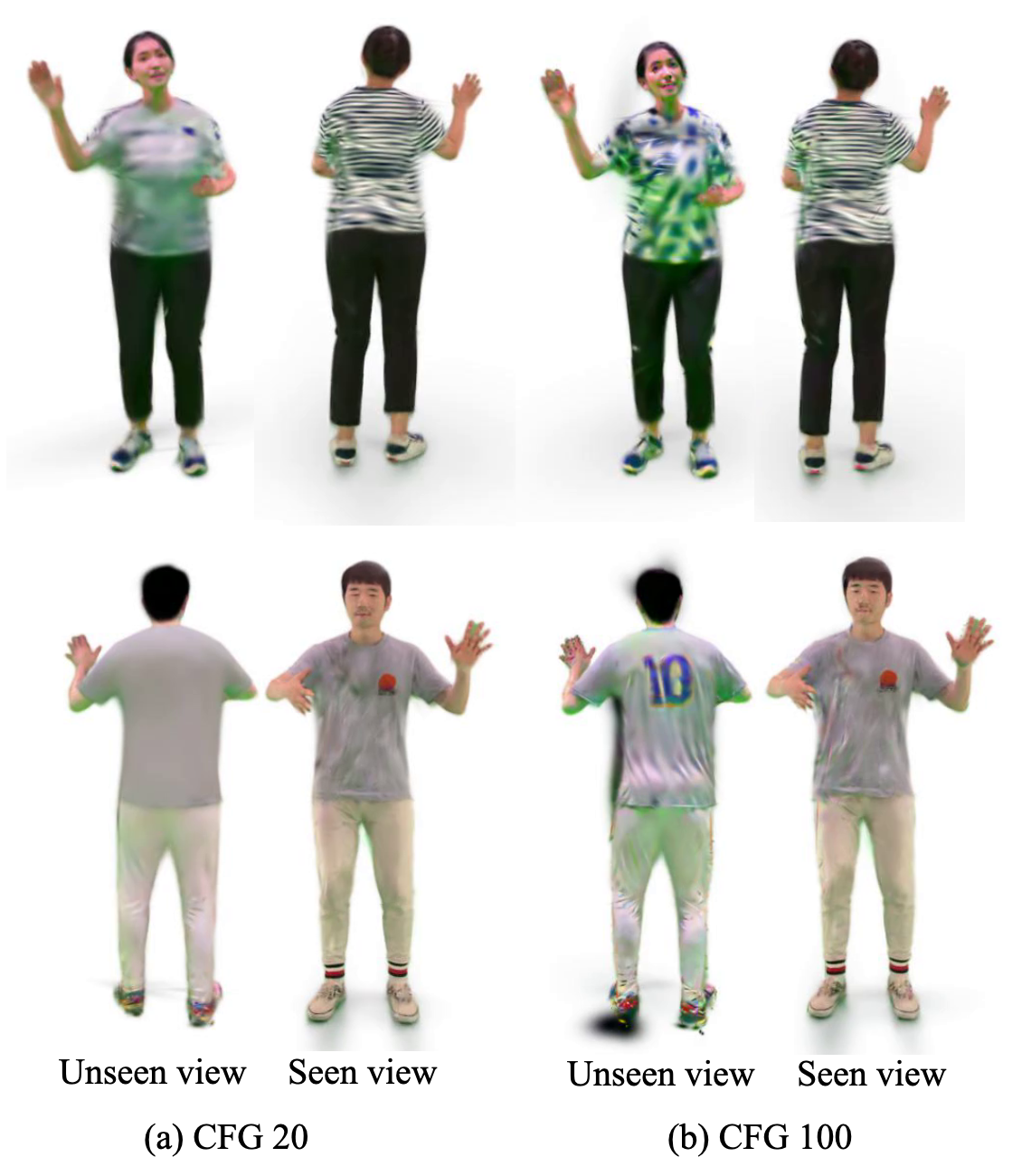}
     \vspace{-20pt}
    \caption{\textbf{Ablation study for the classifier-free guidance scale.} We cropped out the black blurry artifacts near the feet due to lack of space. We can check that a low CFG scale (a) generates a smooth monotonic texture in unseen parts while a high CFG scale (b) synthesizes and enhances wrinkles of clothing on both seen and unseen parts (lower row), but also introduces more artifacts. (upper row)}
    \label{fig:cfg_ablate}
\vspace{-5pt}
\end{figure}

\section{Dataset Preprocessing}
\subsection{Panoptic Dataset~\cite{panoptic_tpami}} 
 We trim the last round of the ultimatum 160422 sequence, extracting 540 multi-view images of 6 individuals by subsampling every 4 frames. Among the 31 HD cameras in the Panoptic Dome, we specifically choose cameras 0, 3, 5, 8, 22, 24, and 25 for evaluation, while camera 16 serves as the input. To simulate a challenging scenario, we intentionally pick the input view camera that excludes the entrance of the Panoptic Dome~\cite{panoptic_tpami} where individuals enter one by one.

 To acquire the SMPL parameters $\vec{\theta}_{t,j}$ and $\vec{\beta}_j$ of individuals, we optimize them by minimizing the distance between 3D SMPL joints and provided pseudo ground truth COCO 3D joints. Our optimization process incorporates pose prior, angle shape regularization, and 3D joint error, as outlined in~\cite{bogo2016smplify}.
  We leverage SMPL joints and SAM~\cite{kirillov2023segment} to obtain each individual's mask in the input frames. Initially, we arrange individuals based on their depth which is calculated as the distance between the pelvis of SMPL and the camera center. Starting with the individual closest to the camera, we obtain a mask by querying the projected SMPL joints which is not occluded into SAM~\cite{kirillov2023segment}. We assume the joints is occluded if it's projected on the masks of nearer people.

\subsection{In-the-wild Videos}
 In handling in-the-wild videos, we categorize them into two scenarios: static camera and moving camera. For the camera moving cases, we employ DROID-SLAM~\cite{teed2021droidslam} to estimate the initial camera pose and Goel et al.~\cite{goel2023humans_in_4d} to track people with regressing SMPL parameters. Subsequently, we refine the estimated parameters by minimizing the reprojection error between estimated 2D body joints~\cite{xu2022vitpose}. In cases with a static camera, we skip the camera pose estimation step.

\begin{figure}[t!]
    \centering
     \includegraphics[width=1.0\columnwidth, trim={6.2cm 0 6.5cm 0},clip]{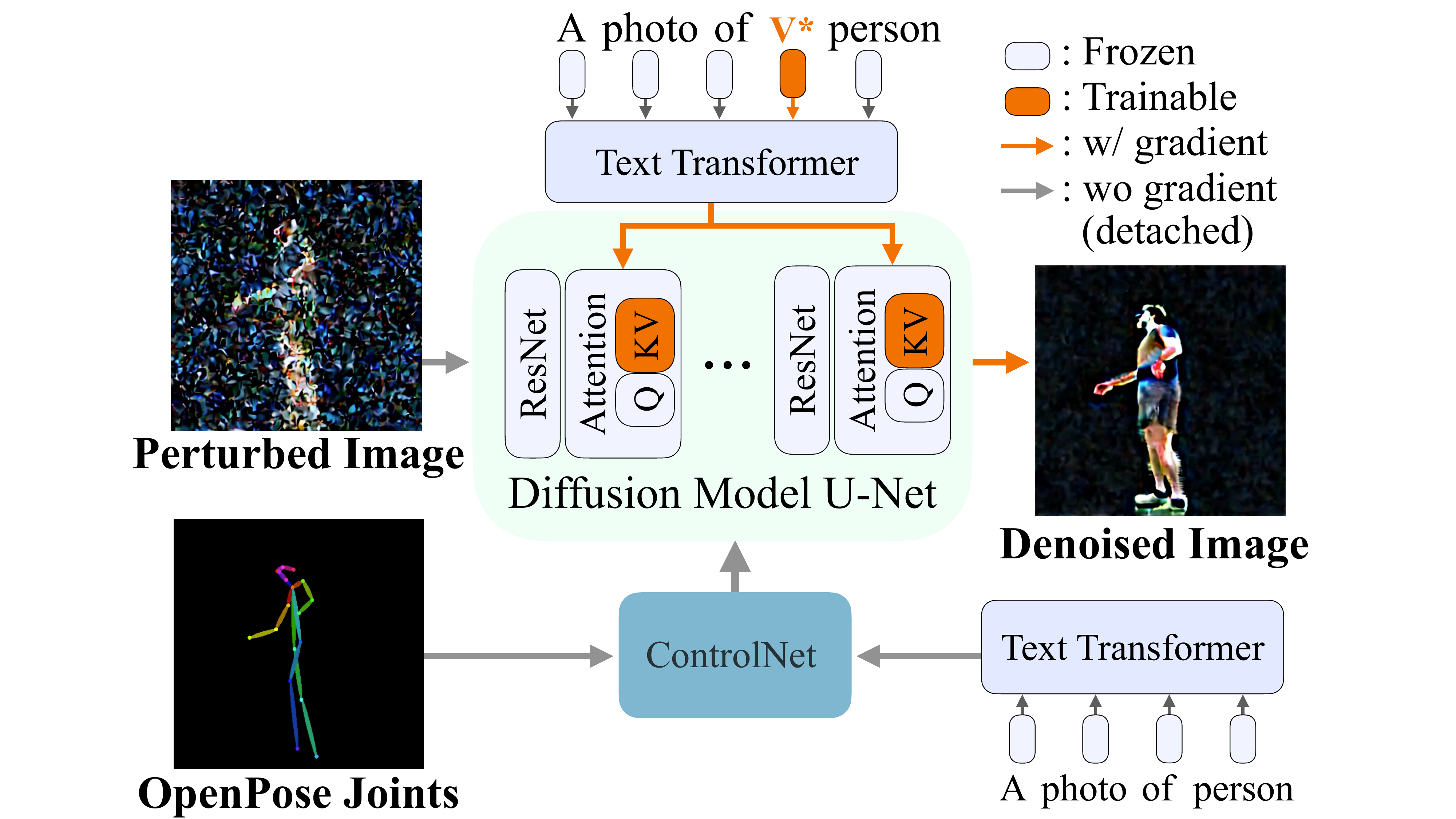}
    \caption{\textbf{Overall Pipeline of Textual Inversion in our method} The orange part is what we optimize during textual inversion. \textbf{V*} indicates textual inversion token \texttt{<person-j>} which is training target. As shown here, we use CustomDiffusion~\cite{kumari2022customdiffusion} together with ControlNet~\cite{zhang2023controlnet} to obtain individuals' inversion token \texttt{<person-j>} and fine-tuned diffusion model $\phi_j$. }
    \label{fig:ti_overview}
\vspace{-5pt}
\end{figure}

\section{Effect of Classifier-Free Guidance Scale} 
 To explore the impact of changing the classifier-free guidance (CFG) scale, we conduct an ablation study using Hi4D~\cite{yin2023hi4d} \texttt{pair00-dance} sequence. As illustrated in the lower row of \figref{fig:cfg_ablate}, a high CFG scale synthesizes detailed unseen parts such as cloth wrinkles and uniform numbers, while a low CFG scale produces a smooth, monotonic texture without any wrinkles. Notably, a high CFG scale introduces more artifacts such as green stains which are amplified by the light reflected from the floor shown in the upper row of \figref{fig:cfg_ablate}. This study shows the importance of selecting a proper CFG scale to reconstruct a detailed human avatar with minimal artifacts. 

\begin{figure}[t!]
    \centering
     \includegraphics[width=1.0\columnwidth, trim={0cm 0 0cm 0},clip]{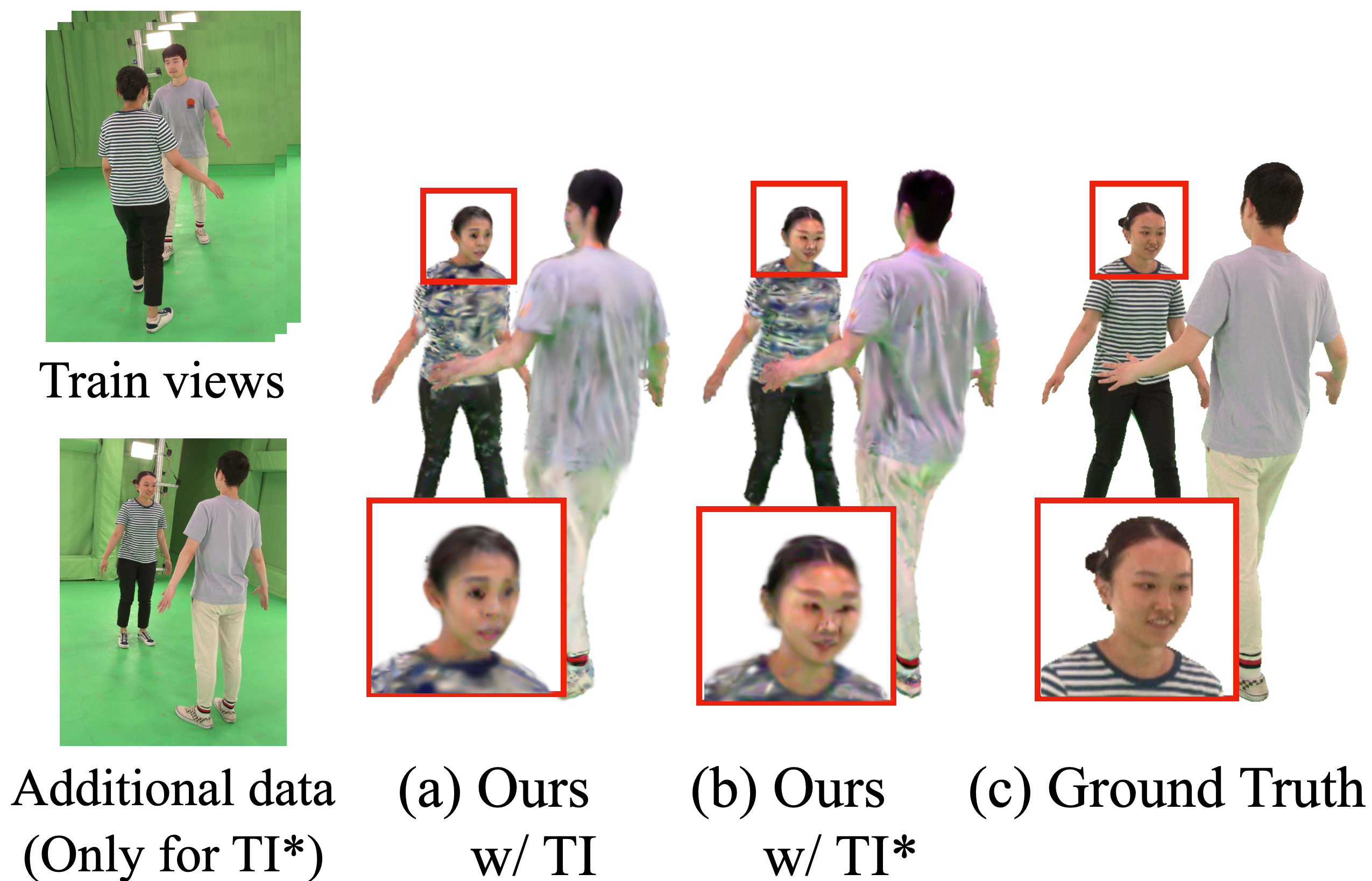}
     \vspace{-10pt}
    \caption{\textbf{Ablation study of adding additional data during Textual Inversion.} TI* means the textual inversion used in SDS loss is trained with a single additional image of the frontal view. Both (a) and (b) are optimized with train views and the only difference is in the Textual Inversion.}
    \label{fig:add_partial_data}
\vspace{-5pt}
\end{figure}

\section{Details of Textual Inversion}
 To obtain an individual's text-token \texttt{<person-j>} and specified fine-tuned diffusion, we run CustomDiffusion on each individual's observations with modifications as shown in \figref{fig:ti_overview}. 
 We use OpenPose ControlNet~\cite{zhang2023controlnet} during Textual Inversion to avoid possible overfitting on observed body pose and camera pose. 
  To obtain an individual's text-token \texttt{<person-j>} and specified fine-tuned diffusion, we first randomly perturb the observed image and then estimate the added noise of the perturbed image.
 By minimizing the MSE loss between the added noise and the estimated noise, we optimize the text-token and fine-tune the diffusion model. 
 As we use the latent diffusion model~\cite{rombach2022stablediffusion_cvpr} here, the training objective is as follows:
  \begin{equation}
     \mathcal{L}_{textual} = \text{MSE}(\epsilon_{\phi}(\vec{z}_\tau; \vec{y}, \tau) - \epsilon)
 \end{equation},
 where $\vec{z}_\tau$ is a perturbed latent corresponding to perturbed image in \figref{fig:ti_overview} and $\epsilon$ is the added noise. 
 During optimization, we randomly sample $\tau$ from $\tau \sim \mathcal{U}[0, 1]$.
 
 We optimize textual token and fine-tune diffusion using Adam~\cite{kingma2014adam} optimizer with learning rate $5e^{-6}$ and batch size 4 for 1000 iterations. 
 To mitigate the situation where the text token learns the background, we mask out the background and randomly fill it with random color. 
 We do not use prior preservation loss here to overfit the text token on observed images. 
 The text-token \texttt{<person-j>} is queried only in Diffusion U-Net and not queried in the ControlNet module as shown in \figref{fig:ti_overview}.

\section{Enhancing Identity with Additional Images}
By employing additional image sources for the target identity, if they are known in advance, we can enhance the identity of the person with sparse observations. 
Specifically, training the Textual Inversion (TI) with an extra face image of the target person, assuming this information is available beforehand, enables our method to produce results that more closely resemble the target human, even in scenarios with an extreme lack of frontal train views. 
We further show such scenario in \figref{fig:add_partial_data} (b), where training the TI with just a single additional frontal image substantially improves the resemblance of the outputs, compared to \figref{fig:add_partial_data} (a). 
This demonstrates the unique advantage of using textual inversion for reconstruction, a method that is difficult to leverage using only reconstruction loss.

\newpage
 \begin{table*}
\caption{Table of notations.}
\centering
\makebox[\textwidth]{
\begin{tabular}{lll}
\toprule
 Symbol  & Description \\
 \midrule
 \multicolumn{2}{c}{\bf Index}\\
 $i$      & Gaussian index, $i\in \{1,\dots, N\}$ in 3D Gaussian attributes\\
 $j$        &   Human index, in human Gaussians $\mathcal{G}^h_j$ and SMPL parameters $\theta_{j,t}, \beta_j$\\
 $t$        &   Time index, $t\in \{1,\dots, T\}$ in SMPL pose parameters, input images \\
 $k$        &   Joint index, $k \in \{1,\dots, N_{joint}\}$ in LBS skinning \\

 \midrule
 \multicolumn{2}{c}{\bf Learnable Attributes of 3D Gaussians}\\
 $\vec{\mu}_i\in \mathbb{R}^3$& Center of $i$-th Gaussian \\
 $\vec{q}_i\in SO(3)$  & Covariance Matrix's Quaternion Component of $i$-th Gaussian \\
 $\vec{s}_i\in \mathbb{R}^3$  & Covariance Matrix's Scale Component of $i$-th Gaussian \\
 $\vec{c}_i\in \mathbb{R}^3$  & Color of $i$-th Gaussian \\
 $o_i\in \mathbb{R}$  & Opacity of $i$-th Gaussian \\
 $G_i$  & $i$-th Gaussian consists of $\{\vec{\mu}_i,  \vec{q}_i,  \vec{s}_i, \vec{c}_i, o_i\}$\\
 \midrule
 \multicolumn{2}{c}{\bf Parameters of Diffusion Model}\\
 $\phi / \phi_j$ & Diffusion model / Diffusion model fine-tuned on $j$-th person \\
 $\tau$ & noise time-step of diffusion model $\tau \in [0,1]$\\
 $\vec{z}_0$ & Encoded latent of the queried RGB images on diffusion model\\
 $\vec{z}_\tau$ & Perturbed latent with noise time-step $\tau \in [0,1]$\\
 $\epsilon$ & Noise added to the latent\\
 $\epsilon_{\phi}$ & Noise estimated by diffusion model $\phi$\\
 \midrule
 \multicolumn{2}{c}{\bf Parameters of Human Deformation}\\
 $\vec{\theta}_{j,t} \in \mathbb{R}^{72} $ & SMPL pose parameter of $j$-th Human in time $t \in \{1,\dots, T\}$ \\
 $\vec{\beta}_{j} \in \mathbb{R}^{10} $ & SMPL shape parameter of $j$-th Human \\
 $\vec{\theta}_{c} \in \mathbb{R}^{72} $ & Canonical pose parameter shared for all humans \\
 \midrule
 \multicolumn{2}{c}{\bf Rendered and Observed Images}\\
 $R_t / I_t$      &   Rendered / Observed RGB image in time $t \in \{1,\dots, T\}$ \\
 $R^h_v$      & Rendered RGB image of a human with camera $v$\\
 \bottomrule
\label{tab:notations}
\end{tabular}
}
\end{table*}
 \fi

\end{document}